\documentclass[times,twocolumn,review]{elsarticle}

\usepackage{medima}
\usepackage{framed,multirow}

\usepackage{amssymb}
\usepackage{latexsym}

\usepackage[export]{adjustbox}
\usepackage{mathtools}
\usepackage{bm}
\usepackage{subcaption}
\usepackage{caption}

\usepackage{times}
\usepackage{epsfig}
\usepackage{graphicx}
\usepackage{amsmath}
\usepackage{multirow}
\usepackage{xcolor}
\usepackage[multi-part-units=single]{siunitx}
\usepackage{lineno}

\usepackage{url}
\usepackage{xcolor}

\usepackage{hyperref}

\definecolor{newcolor}{rgb}{.8,.349,.1}

\journal{Medical Image Analysis}

\newcommand*{\myDots}{\ifmmode.\kern-0.13em.\kern-0.13em.\else.\kern-0.13em.\kern-0.13em.\fi}

\begin{document}

\verso{Nils Gessert \textit{et~al.}}

\begin{frontmatter}

\title{Deep learning with 4D spatio-temporal data representations for OCT-based force estimation}%

\author[1]{Nils \snm{Gessert}\corref{cor1}\fnref{fn1}}
\ead{nils.gessert@tuhh.de}
\cortext[cor1]{Corresponding author}
\author[1]{Marcel \snm{Bengs}\fnref{fn1}}
\fntext[fn1]{Authors contributed equally.}
\author[1]{Matthias \snm{Schl\"uter}}
\author[1]{Alexander \snm{Schlaefer}}

\address[1]{Hamburg University of Technology, Institute of Medical Technology, Am Schwarzenberg-Campus 3, 21073 Hamburg}

\received{15 November 2019}
\finalform{20 May 2020}
\accepted{20 May 2020}
\availableonline{20 May 2020}
\communicated{TBD}

\begin{abstract}
Estimating the forces acting between instruments and tissue is a challenging problem for robot-assisted minimally-invasive surgery. Recently, numerous vision-based methods have been proposed to replace electro-mechanical approaches.
Moreover, optical coherence tomography (OCT) and deep learning have been used for estimating forces based on deformation observed in volumetric image data. The method demonstrated the advantage of deep learning with 3D volumetric data over 2D depth images for force estimation. 
In this work, we extend the problem of deep learning-based force estimation to 4D spatio-temporal data with streams of 3D OCT volumes. For this purpose, we design and evaluate several methods extending spatio-temporal deep learning to 4D which is largely unexplored so far. Furthermore, we provide an in-depth analysis of multi-dimensional image data representations for force estimation, comparing our 4D approach to previous, lower-dimensional methods. Also, we analyze the effect of temporal information and we study the prediction of short-term future force values, which could facilitate safety features. For our 4D force estimation architectures, we find that efficient decoupling of spatial and temporal processing is advantageous. We show that using 4D spatio-temporal data outperforms all previously used data representations with a mean absolute error of $\SI{10.7}{\milli\newton}$. We find that temporal information is valuable for force estimation and we demonstrate the feasibility of force prediction.
\end{abstract}


\end{frontmatter}


\section{Introduction}

For minimally invasive interventions, the use of robotic assistance systems can provide tremor compensation or motion scaling for reduced physical trauma \citep{kroh2015essentials}. For these systems force measurement between instruments and tissue can be beneficial for haptic feedback or to implement safety measures stopping the robotic instrument when forces exceed some threshold \citep{haidegger2009force}. Previously, electro-mechanical force sensors have been directly integrated into the surgical setup which is associated with challenges such as sensor size and sterilization \citep{sokhanvar2012clinical}. Therefore, vision-based force estimation has been proposed where forces are directly learned from RGB(D) images of the deformed tissue \citep{greminger2004vision}. 

Early methods for vision-based force estimation relied on deformable template matching methods \citep{greminger2004vision}. Also, \cite{greminger2003modeling} proposed to learn forces based on handcrafted features and conventional machine learning methods. More recently, temporal information has been incorporated into force estimation models. This follows the idea of tracking tissue deformation over time which provides a more reliable estimate than a single-shot measurement as tissue is often in continuous motion during surgery \citep{aviles2015force}.  \cite{marban2019recurrent} extended this concept with a convolutional neural network (CNN) for spatial image processing. As an alternative to RGBD images, optical coherence tomography (OCT) has been proposed as an imaging modality for force estimation \citep{otte2016towards}. OCT can provide 3D volumes with a rich spatial feature space through subsurface imaging which can be effectively used by deep learning methods \citep{gessert2018deep}. OCT's ability to capture subsurface tissue compression has been used with 3D CNNs that estimate forces from single 3D volumes \citep{gessert2018force}. Summarized, recent approaches for vision-based force estimation either relied on 3D spatio-temporal data (streams of RGBD images) or 3D spatial data (OCT volumes). 

\begin{figure*}[t]
\begin{center}
   \includegraphics[width=1.0\linewidth]{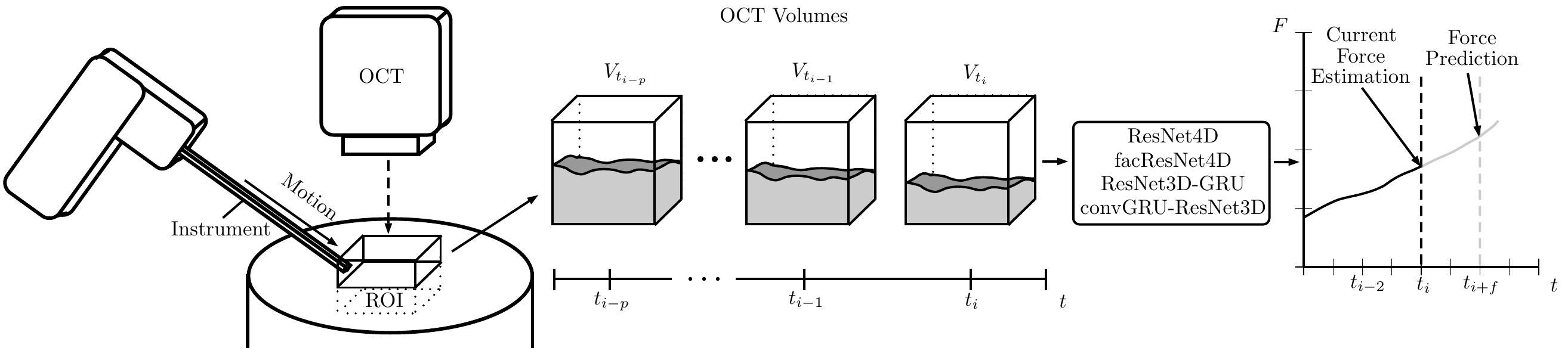}
\end{center}
   \caption{Illustration of the concept of force estimation from a series of OCT volumes. Here, tissue or material is deformed by an instrument that is captured by a series of OCT volumes. Then, one of our proposed deep learning models estimates or predicts a force using the 4D data stream.}
\label{fig:concept}
\end{figure*}

In this paper, we extend vision-based force estimation to 4D using streams of 3D OCT volumes. This requires extending spatio-temporal deep learning approaches to 4D data. So far, spatio-temporal deep learning models have been largely employed for 3D spatio-temporal data, usually streams of 2D images or depth images over time. Most spatio-temporal models emerged from the natural image domain for tasks such as human action recognition \citep{simonyan2014two} and video classification \citep{tran2015learning}. Typical approaches include 2D CNNs with recurrent architectures \citep{donahue2015long}, 3D CNNs with full temporal convolutions \citep{ji20133d}, factorized temporal convolutions \citep{sun2015human} or two-stream architectures with a spatial and a temporal data stream \citep{simonyan2014two}.

Time series of volumes represent 4D data and are uncommon outside of medical applications. Thus, there are only a few deep learning approaches with 4D data \citep{choy20194d}. In the medical image domain, 4D CNNs have been used with 4D computed tomography (CT) \citep{clark2019convolutional,myronenko20194d}. The approaches showed the feasibility of applying 4D CNNs, however, no performance improvement over 3D CNNs was presented. Moreover, other approaches have used fused recurrent and convolutional models with 4D magnetic resonance imaging (MRI) and CT data \citep{zhao2018modeling,van2019stacked,bengs2019}. While showing promising results, these approaches generally lack an extensive evaluation of different methods for processing spatial and temporal data dimensions.

So far, there is no work on 4D deep learning using OCT data and 4D deep learning for vision-based force estimation. We hypothesize that high-dimensional 4D spatio-temporal OCT data could provide additional information and therefore improve performance. However, several open challenges need to be addressed when dealing with 4D spatio-temporal OCT data.

The first challenge we address is architecture design for 4D spatio-temporal deep learning. This task is very difficult due to the substantial increase in computational and memory requirements compared to architectures for lower-dimensional data. Also, it is unclear what type of spatial and temporal processing mechanisms should be employed. While a lot of spatio-temporal deep learning methods have been presented for lower-dimensional 3D data, extensions to 4D are limited. In particular, previous 4D approaches \citep{zhao2018modeling,van2019stacked,clark2019convolutional,myronenko20194d,choy20194d} lack a systematic comparison of different spatio-temporal processing mechanisms and often only rely on a single architecture concept. Therefore, we design and evaluate four very different approaches for efficiently processing the spatial and temporal data dimensions of 4D spatio-temporal OCT data.

The second challenge we address is the high-dimensional nature of OCT data. While  4D spatio-temporal data should be rich in information, processing it is computationally expensive. One approach for reducing computational effort is to encode the spatial 3D volume information in a lower-dimensional space. For vision-based force estimation, it is reasonable to assume that the tissue \textit{surface} being deformed by a force carries information that can be encoded in a space that is smaller than full 3D volumes. For example, previous vision-based force estimation methods have captured deformation with a feature vector or 2D depth maps \citep{aviles2015force}. This raises the question of whether full 3D volumes are advantageous or whether lower-dimensional surface encodings are sufficient. Therefore, we study different representations of the spatial image information by considering full 3D volumes as well as 2D and 3D image data representations of the deformed surface.

Another way of controlling 4D spatio-temporal data processing effort is to reduce or extend the temporal dimension. Thus, the third challenge we address is the effective use of the temporal data dimension. We investigate and quantify the effect and benefit of different temporal processing techniques as well as a longer or shorter temporal history. In this context, we also consider the problem of short-term force \textit{prediction}. Assuming some non-random deformation and force patterns over time, forces should be predictable, given a history of data. Predicting forces could enable safety mechanisms for robot-assisted interventions \citep{haidegger2009force,haouchine2018vision} as large force value increases could be detected earlier. Although spatio-temporal data has been used for force estimation, prediction has not been studied.

The main contributions of this paper are three-fold. First, we present four different 4D spatio-temporal deep learning methods and evaluate them for OCT-based force estimation. Second, we demonstrate that 4D deep learning outperforms previous approaches using lower-dimensional data representations. Third, we evaluate the effect of the temporal dimension and demonstrate the feasibility of short-term force prediction. To facilitate further research on 4D deep learning we make our code publicly available\footnote{https://github.com/ngessert/4d\_deep\_learning}.

\section{Related Work} \label{sec:related_work}

In this section, we consider the relevant literature for our work. Our approach is related to the fields of vision-based force estimation, spatio-temporal deep learning and 4D deep learning.

\textbf{Vision-based force estimation} is the task of estimating forces that are acting on tissue, only based on images that are typically provided by RGB-D cameras. This avoids integrating mechanical force sensors into surgical setups which is often difficult and brings up issues such as sterilization and biocompatibility \citep{sokhanvar2012clinical}. Vision-based force estimation was initially not considered a spatio-temporal data processing problem as early methods relied on deformable template matching methods \citep{greminger2004vision}. Similarly, following methods relied on mechanical deformation models \citep{kim2010haptic,kim2012image,noohi2014using}. Furthermore, \cite{greminger2003modeling} proposed to learn forces based on handcrafted features which found wider adoption \citep{karimirad2014vision,mozaffari2014identifying}. More recently, the temporal dimension has been integrated into force estimation models by using handcrafted features derived from stereoscopic camera images in LSTM models \citep{aviles2015force,aviles2017towards}. This follows the idea of tracking tissue deformation over time which provides a more reliable estimate than a single-shot measurement during surgery where the tissue is in continuous motion. Furthermore, taking past deformation into account allows for force \textit{prediction} which could enable safety mechanisms for robot-assisted interventions \citep{haidegger2009force,haouchine2018vision} but has not been addressed so far. Spatio-temporal approaches were extended by \cite{marban2019recurrent} where a 2D CNN extracts spatial features that are then fed into an LSTM. \cite{gao2018learning} use both a 2D CNN to learn spatial features from RGB images as well as PointNet \citep{qi2017pointnet} to process point clouds derived from depth images. Features from both networks are fed into an additional CNN for temporal processing. Thus, the approach takes 3D information into account by using depth, however, the explicit 4D problem is avoided.
While most methods rely on RGB and depth cameras, \cite{otte2016towards} proposed to use OCT as an imaging modality. Recently, \cite{gessert2018force} showed that exploiting the subsurface imaging capabilities of OCT with 3D volumes leads to improvements over the use of 2D depth maps only. This study performed force estimation in a single-shot style. For needle insertion scenarios, OCT-based force estimation has also been studied as a spatio-temporal learning problem with 1D images over time \citep{gessert2019spatio}. Summarized, force estimation has been addressed with different 2D and 3D image data representation. A concise comparison of multi-dimensional data representations is still missing and 4D data has not been addressed at all. While temporal information has been considered, the task of short-term force prediction has not been considered. 

\textbf{Spatio-temporal deep learning methods }mostly differ in the way the temporal dimension is treated and commonly rely on CNNs to treat the spatial dimensions \citep{asadi2017survey}. One class of models uses 3D CNNs with kernels $K \in \mathbb{R}^{k_t\times k_h\times k_w\times k_c}$ to incorporate the temporal dimension \citep{liu20163d}. \cite{ji20133d} are amongst the early adopters of 3D CNNs for human action recognition. \cite{tran2015learning} investigated the use of 3D CNNs further on large-scale datasets and proposed the widely adopted C3D architecture. Recently, \cite{varol2018long} studied long-term convolutions with different lengths of sequences using 3D CNNs. As 3D convolutions lead to more model parameters, more efficient approaches have been proposed. \cite{sun2015human} proposed factorized convolutions where the 3D kernel is split into a 2D spatial and 1D temporal kernel that are applied sequentially to the data. \cite{qiu2017learning} extend this approach by investigating different variants of residual blocks \citep{he2016identity} for separate spatial and temporal convolutions. Recently, \cite{tran2018closer} proposed a 3D CNN with mixed convolutional layers where 3D convolutions are only applied in lower layers of the model. Using convolutions for temporal processing has been applied in the medical domain, e.g., for surgical video analysis \citep{funke2019using}. 

Another class of architectures attempts to model temporal relations with recurrent architectures. Commonly, a 2D CNN performs spatial processing first which is followed by a recurrent part \citep{ordonez2016deep}. \cite{donahue2015long} and \cite{yue2015beyond} used this concept by feeding features from a 2D CNN into LSTM \citep{hochreiter1997long} layers. \cite{pigou2018beyond} investigated different variants of spatio-temporal models including CNN+LSTM models and temporal pooling. This has also been applied in the medical domain, e.g., for surgical video analysis \citep{jin2019multi} and force estimation \citep{gao2018learning}.
Also, two-stream architectures have been proposed where one convolutional path receives spatial information in the form of single frames and one convolutional path receives temporal information, e.g., in the form of precomputed optical flow \citep{simonyan2014two,feichtenhofer2016convolutional,wang2016temporal}. 

Overall, multiple very different ways of processing spatial and temporal data have been proposed. To provide a comprehensive analysis, we consider four different architectures for spatio-temporal data processing, relying on both convolutional and recurrent concepts.

\textbf{4D deep learning models} for 4D spatio-temporal data are still rare. In the natural image domain, 3D spatial information obtained from time-of-flight cameras can often be encoded as a depth map which avoids full 4D data processing. \citep{el2018yolo4d} map sequences of 3D LiDAR point clouds into 2D projections that are fed into convolutional LSTMs \citep{xingjian2015convolutional} for concurrent spatial and temporal processing. \cite{choy20194d} propose sparse 4D convolutions for directly processing 4D spatio-temporal data from depth sensors. In the medical imaging domain, 4D CNN models have been used for CT image reconstruction \citep{clark2019convolutional} and segmentation \citep{myronenko20194d}. While showing the feasibility of apply 4D deep learning, these approaches do not show an improvement over the use of lower-dimensional and thus often more efficient, deep learning methods. Also, image-to-image translation \citep{van2019stacked}, functional MRI (fMRI) modeling \citep{zhao2018modeling} and fMRI-based disease classification \citep{bengs2019} have been shown. 

Summarized, there have been no 4D deep learning studies for OCT data or any force estimation approach. Moreover, we observe that 4D deep learning studies tend to rely on a few or only one concept for spatio-temporal data processing. This motivates our approach of considering multiple different spatio-temporal deep learning approaches. 

\section{Methods} \label{sec:methods}

\subsection{Problem Definition and Data Representations}

\begin{figure}[ht]
\begin{center}
   \includegraphics[width=1.0\linewidth]{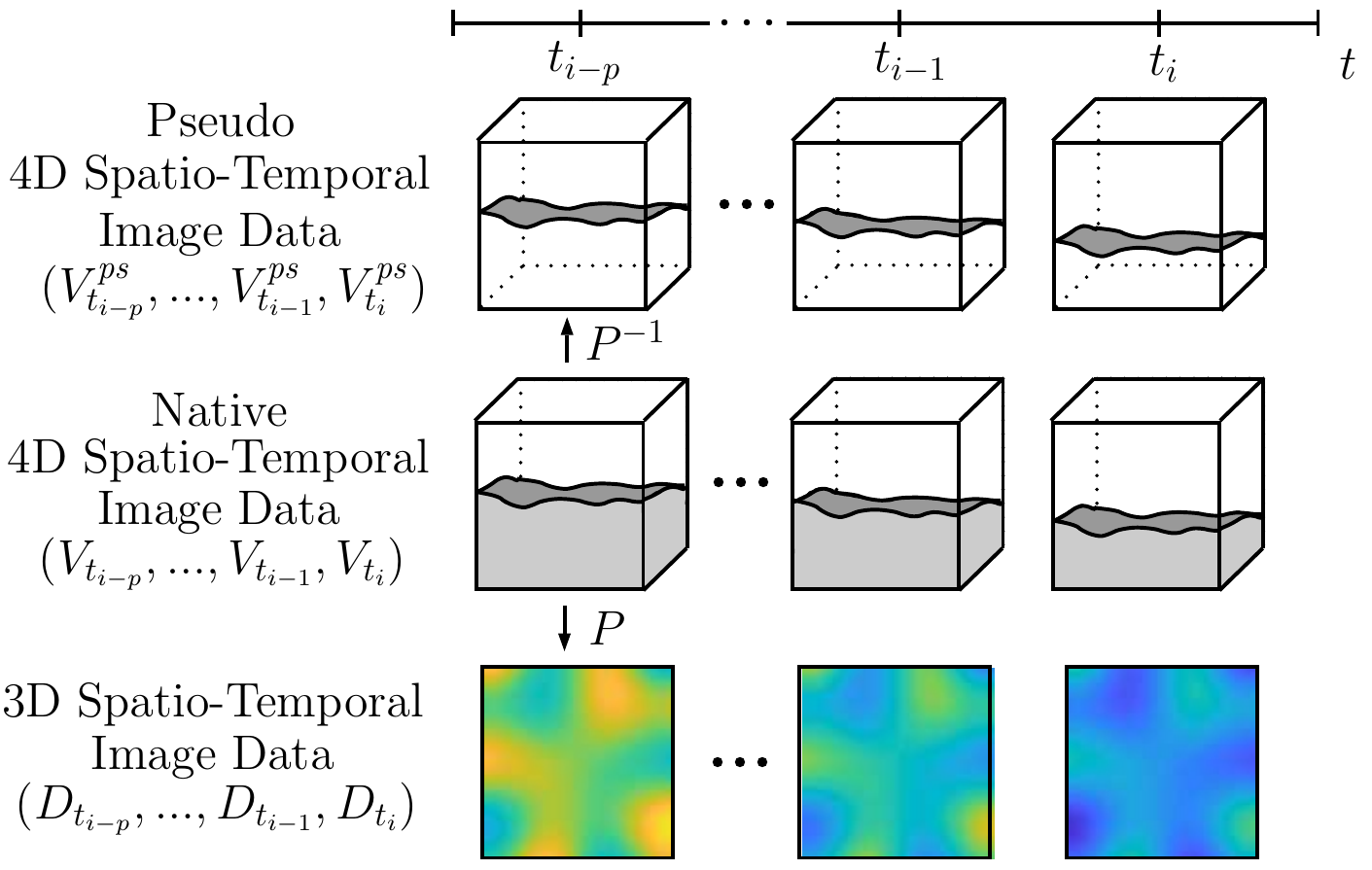}
\end{center}
   \caption{The image data representations that we study. In the middle row, a sequence of full volumes is shown. Bottom, a sequence of the extracted surface encoded as depth images is shown. Depth is represented by color value. Top, the extracted surfaces are represented by points in a volume.} 
\label{fig:reps}
\end{figure}

Formally, we consider a sequence of volumes $(V_{t_{i-p}},...,V_{t_{i-1}},V_{t_i})$ which is used to estimate the force $F_{t_{i+f}}$. The volumes $V_{t_i} \in \mathbb{R}^{h\times w\times d}$ capture the deformation of tissue to which forces $F_{t_i} \in \mathbb{R}$ are applied by an instrument. Along the temporal dimension, $p$ denotes the history, i.e., the number of past volumes used for prediction, and $f$ denotes the prediction horizion, i.e., at how many time steps in the future we predict. For $f=0$ we refer to the problem as force \textit{estimation}, for $f>0$ we refer to the problem as force \textit{prediction}.

We also study two transformations of spatial data. First, we consider a projection $P:\mathbb{R}^{h\times w\times d}\to\mathbb{R}^{h\times w}$. The resulting depth images $D_{t_i} \in \mathbb{R}^{h\times w}$ represent a surface within a 3D volume. Second, we represent the surface by points in the actual 3D volume, i.e., we consider $P^{-1}:\mathbb{R}^{h\times w}\to\mathbb{R}^{h\times w\times d}$ where depth images are encoded as \textit{pseudo volumes} $V_{t_i}^{\mathit{ps}} \in \mathbb{R}^{h\times w\times d}$ in 3D space. All data representations are shown in Figure~\ref{fig:reps}.

We design and evaluate models with spatio-temporal (ST) and spatial (S) data for learning $M:\mathbb{R}^{p\times h\times w\times d}\to\mathbb{R}$ (4D-ST) and compare to models $M:\mathbb{R}^{p\times h\times w\times d}\to\mathbb{R}$ (ps-4D-ST), $M:\mathbb{R}^{p\times h\times w}\to\mathbb{R}$ (3D-ST), $M:\mathbb{R}^{h\times w\times d}\to\mathbb{R}$ (3D-S) and $M:\mathbb{R}^{h\times w}\to\mathbb{R}$ (2D-S). 

\subsection{Deep Learning Architectures for 4D Data}

\begin{figure*}[ht]
\begin{center}
   \includegraphics[width=1.0\linewidth]{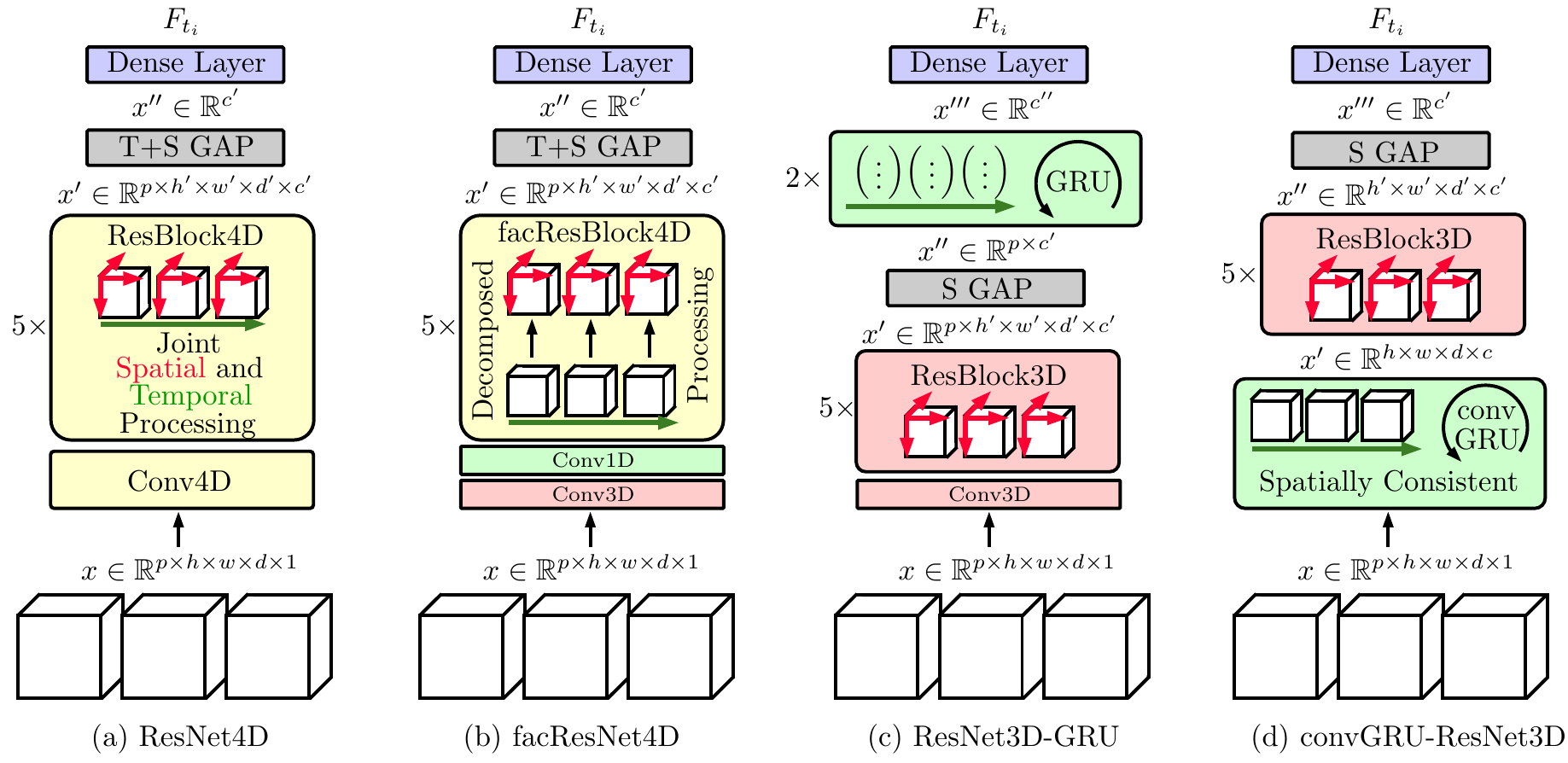}
\end{center}
   \caption{The network architectures we propose. Each architecture receives a sequence of volumes as its input. T+S GAP refers to global average pooling over the temporal and spatial dimensions. Red indicates spatial processing, green indicates temporal processing and yellow indicates joint processing.} 
\label{fig:arch_overview}
\end{figure*}

In this section, we describe our 4D-ST architectures shown in Fig.~\ref{fig:arch_overview}. For each model, we consider a 3D-ST version for comparison that matches the 4D-ST versions in terms of structure. The CNN part of each architecture is built on the ResNet principle \citep{he2016deep}. We use vanilla ResNet blocks without bottlenecks or other variations. While building upon the same ResNet backbone, each architecture uses a very different way of processing the temporal dimension together with the spatial dimension. 

\textbf{ResNet4D} is an extension of the ResNet principle to 4D convolutions. The mathematical extension of a discrete convolution to $N$ dimensions can be described as 

\begin{flalign}
\begin{aligned}
	(K*x)(j_1,\myDots,j_N) \coloneqq & \\ \sum_{k_1}\myDots \sum_{k_N}K(k_1,\myDots,k_N) & x(j_1-k_1,\myDots,j_N-k_N) .
\end{aligned}
\end{flalign}

In a 4D convolutional layer, the kernel $K^{(l)} \in \mathbb{R}^{k_t\times k_h\times k_w\times k_d \times k_c}$ of layer $l$ is applied to feature maps $x^{(l-1)} \in \mathbb{R}^{p\times h\times w\times d\times c}$, excluding the batch dimension. $p$ is the size of the temporal dimension, $h$, $w$, and $d$ are the spatial extent of the feature map and $c$ is the channel dimension of the feature map. Currently, there are no native 4D convolution operations available in standard environments such as PyTorch and Tensorflow. To keep the 4D convolution as efficient as possible, we implement a custom version in Tensorflow which uses the native 3D convolution operation inside of two loops. The operation can be described as

\begin{equation}
	(K^{(l)}*x^{(l-1)}) = \sum_{i}^{k_t}\sum_{j}^{p}\mathit{Conv3D}(K^{(l)}(i),x^{(l-1)}(j))
\end{equation}

with correct padding and a stride of one assumed. 
The final architecture shown in Fig.~\ref{fig:arch_overview} starts with a normal convolutional layer, followed by five ResBlocks with a spatial output stride of $s_o=16$. The initial feature maps size is $c=16$ which is doubled each time we halve the spatial dimension using a convolution with spatial stride $s=2$. As we keep $p$ small, the temporal dimension is not reduced by strides.
 
For models using 3D-ST data, we use the same architecture with 3D convolutions (ResNet3D-ST), i.e., feature maps are of size $x^{(l)} \in \mathbb{R}^{p\times h\times w\times c}$. Naturally, 4D models come with an increase in parameters. Thus, a fair comparison requires an increase in parameters for the 3D models. To match all models' capacity, we also consider deeper (ResNet3D-ST-D) and wider (ResNet3D-ST-W) 3D variants. For the deeper version, we use a total of $\num{9}$ ResBlocks instead of $\num{5}$ in the baseline model. For the wider version, we double the number of feature maps, i.e., we use $c=32$ instead of $c=16$ for the initial convolutional layer. We use kernels of size $k=3$ for every dimension. 

For spatial approaches using 3D-S and 2D-S data, we consider a ResNet3D-S and ResNet2D-S with 3D and 2D convolutional operations, respectively. In terms of structure, the CNNs are similar to the ST models, however, they receive and process spatial information only. Thus, ResNet2D-S produces feature maps of size $x^{(l)} \in \mathbb{R}^{h\times w\times c}$ and ResNet3D-S produces feature maps of size $x^{(l)} \in \mathbb{R}^{h\times w\times d\times c}$. For these models, we also consider model versions with an increased number of parameters to match 4D models' capacity, similar to ResNet3D-ST-D and ResNet3D-ST-W.

\textbf{facResNet4D} is a more efficient variant of ResNet4D which uses factorized convolutions. In each ResBlock, each convolution is split into two convolutions with spatial kernels $K_{S}^{(l)} \in \mathbb{R}^{1\times k_h\times k_w\times k_d\times k_c}$ and temporal kernels $K_{T}^{(l)} \in \mathbb{R}^{k_t \times 1\times 1\times 1\times k_c}$. This modification leads to a reduced number of parameters and decomposes spatial and temporal computations. Thus, the architecture is more efficient but does not have the same representational power as ResNet4D as the decomposed convolutions can only represent separable 4D kernels. 

For 3D-ST data, we employ a facResNet3D model where the spatial kernels are of size $K_{S}^{(l)} \in \mathbb{R}^{1\times k_h\times k_w \times k_c}$, similar to ResNet3D-ST. The temporal kernels stay the same. In terms of implementation, one of the singleton dimensions is removed.

\textbf{ResNet3D-GRU} first performs spatial feature extraction from the individual volumes using a 3D CNN with kernels of size $K_{S}^{(l)} \in \mathbb{R}^{k_h\times k_w\times k_d\times k_c}$. The entire feature maps that are being processed are of size $x^{(l)} \in \mathbb{R}^{p\times h\times w\times d\times c}$ and thus also contain a temporal dimension. However, the temporal dimension remains untouched during initial, spatial processing as all time points are processed independently in the same way. Then, spatial global average pooling is applied which results in a feature vector of size $x^{(l)} \in \mathbb{R}^{p\times c}$. Then, we perform temporal processing of these abstract, spatial representations using two recurrent layers. The recurrent layers use gated recurrent units (GRU) \citep{cho2014learning}, augmented by recurrent batch normalization \citep{cooijmans2016recurrent}. GRUs are a more efficient version of long short-term memory (LSTM) cells \citep{hochreiter1997long} which popularized gating in recurrent neural networks. As GRUs fuse the LSTM's forget and input gate into a single update gate, GRUs are more parameter efficient. Together with the fusion of the LSTM's cell state and hidden state into a single state, GRUs are therefore also more efficient in terms of memory requirements. All these properties are very attractive in the context of 4D deep learning where efficiency is crucial. To draw a comparison to LSTMs, we also consider a ResNet3D-LSTM variant.

In contrast to ResNet4D, this architecture completely separates spatial and temporal data processing. facResNet4D also separates spatial and temporal processing, however, spatial and temporal processing alternates between layers. Despite the separate processing steps of ResNet3D-GRU, the entire architecture is trained end-to-end. 

For 3D-ST data, we design ResNet2D-GRU where the initial ResNet for spatial processing uses 2D convolutions to process the 2D depth images. Thus, the kernels are of size $K_{S}^{(l)} \in \mathbb{R}^{k_h\times k_w\times k_c}$. After pooling the spatial representation into a feature vector, the same temporal processing with two GRU layers is applied.

\textbf{convGRU-ResNet3D} first performs temporal processing with convolutional GRUs which keep the spatial structure intact \citep{gessert2018needle,gessert2019spatio}. The convGRU also employs recurrent batch normalization and the gates use 3D convolutions. The last temporal output of the convGRU is passed to a 3D CNN, i.e., the convGRU's output has a size of $x^{(l)} \in \mathbb{R}^{h\times w\times d\times c}$. The following ResNet3D uses normal 3D convolution operations and performs spatial processing only. Compared to ResNet4D and facResNet4D, this architecture also keeps the temporal and the spatial processing parts separate. The key difference to ResNet3D-GRU is that the order of temporal and spatial processing is reversed. As the spatial structure of the input data needs to be kept intact during temporal processing, convolutions for the gating operations \citep{xingjian2015convolutional} are well suited for this network. In contrast to ResNet3D-GRU, the architecture therefore learns smaller, localized temporal dependencies instead of more global relationships based on abstract feature vectors.

Similar to ResNet3D-GRU, we also consider a variant of this architecture where LSTMs are used instead of GRUs. For this convLSTM-ResNet3D architecture, we make use of convolutional LSTMs which are structurally similar to LSTMs but use 3D convolutional layers instead of matrix multiplications.

For 3D-ST data, we use a convGRU-ResNet2D where all 3D convolution operations are replaced by 2D variants. All other properties are the same as for convGRU-ResNet3D.

\textbf{Training.} For the different model variants, we chose learning rate and batch size individually based on validation performance. For 3D architectures, we use a batch size of $b_s=16$ and learning rate of $l_r = \num{5e-4}$. For 4D architectures we use a batch size of $b_s=8$ and a learning rate of $l_r=\num{2.5e-4}$. During validation experiments, we did not observe any overfitting and the loss converged well. Weights are initialized using a truncated normal distribution with zero mean and standard deviation $s_d=0.01$ where values are redrawn if their magnitude is larger than twice the standard deviation. We train for $100$ epochs using the Adam algorithm with the recommended standard parameters \citep{Kingma.2014}. During training, we track exponential moving averages of all trainable parameters with a decay rate of $e_d = \num{0.999}$. During evaluation, we use the moving average of all model parameters for more consistency compared to a point estimate of parameters at the last iteration. The loss function is the mean squared error. We implement our models and training environment in Tensorflow \citep{tensorflow2015-whitepaper}. Training is performed on NVIDIA GEFORCE 1080Ti graphics cards.

\subsection{Experimental Setup and Datasets}

\begin{figure}[t]
\begin{center}
   		\includegraphics[width=1.0\linewidth]{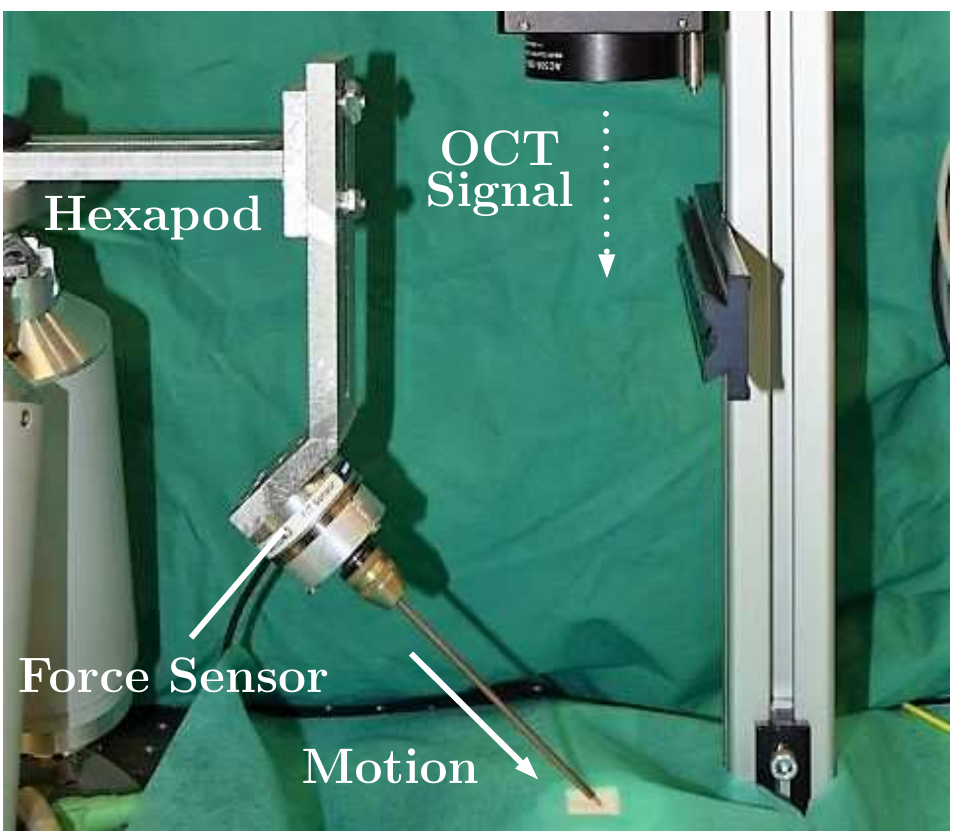}
\end{center}
   \caption{The experimental setup we use for data acquisition.}
\label{fig:exp_setup}
\end{figure}

\textbf{The experimental setup} is shown in Fig.~\ref{fig:exp_setup}. A hexapod robot (H-820.D1, Physik Instrumente) is equipped with a force sensor (Nano 43, ATI) for ground-truth annotation and an instrument. We use a needle tool for our experiments. The hexapod performs movement along the needle axis which deforms a silicon phantom, representing a typical surgical pushing task \citep{marban2019recurrent}. The phantom is imaged by an OCT scan head which continuously acquires volumes. Note that the force sensor is only required for training data acquisition and not for the actual application. To ensure generalization to real-world tissue applications, we also consider a dataset where animal heart muscle tissue is used instead of a phantom.

\begin{figure}[t]
\begin{center}
   \includegraphics[trim=0 0 0 0.5cm, clip, width=1.0\linewidth]{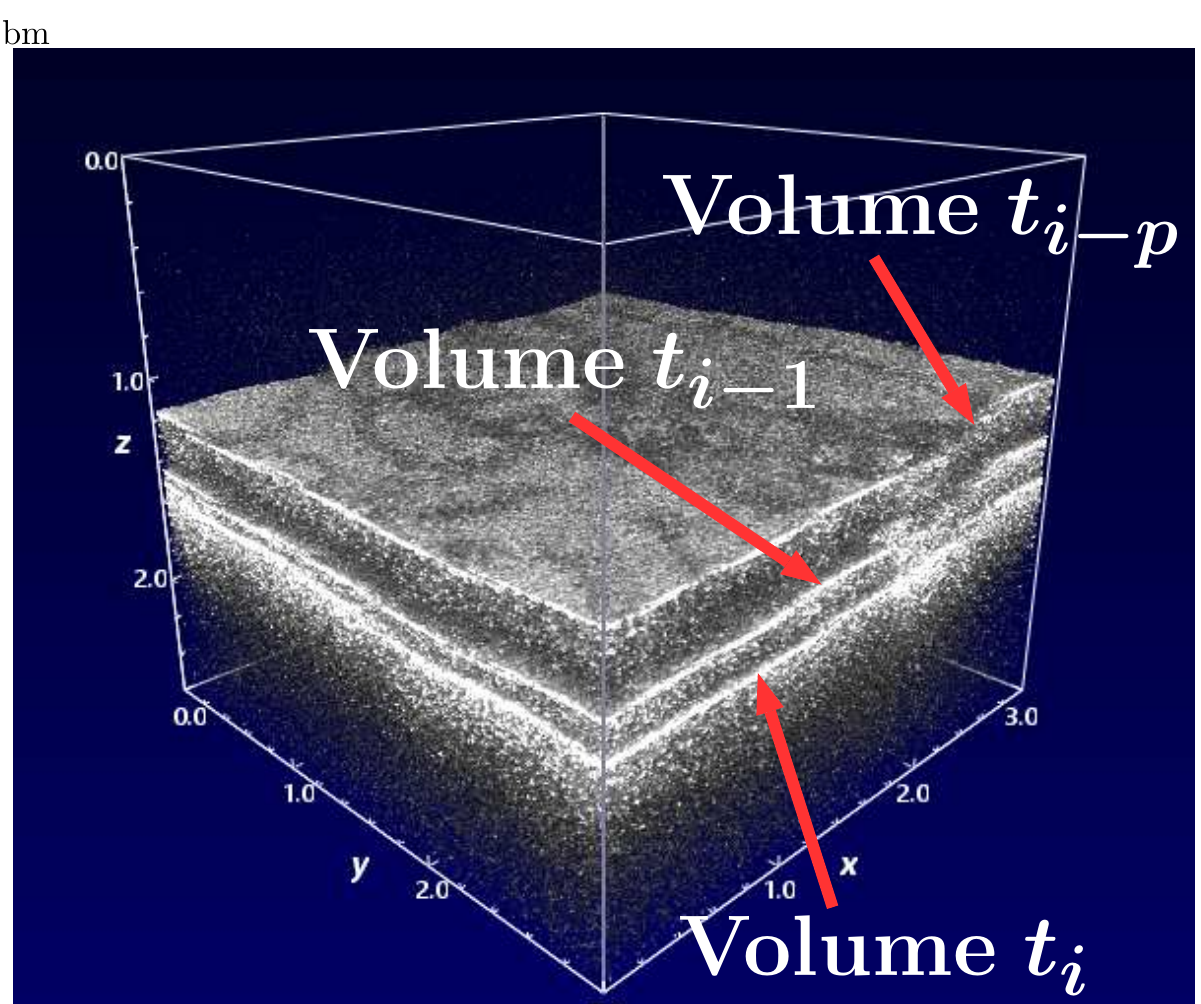}
\end{center}
   \caption{Three overlaid OCT volumes from three time steps.}
\label{fig:example_vols}
\end{figure}

\begin{figure}[t]
\begin{center}
   \includegraphics[width=1.0\linewidth]{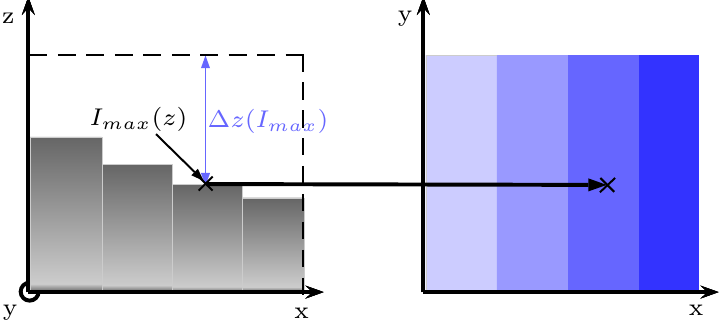}
\end{center}
   \caption{Illustration of projection $P$ where the index of a maximum intensity projection is used to derive a depth map from OCT volumes. Left, an OCT slice is depicted where darker values correspond to high intensities. Right, the projected depth map is shown. Darker shades of blue indicate larger depth values.}
\label{fig:depth_maps}
\end{figure}

\textbf{The OCT device} is a swept-source OCT (OMES, OptoRes, Germany). Based on interferometry and infrared light with a central wavelength of $\SI{1315}{\nano\metre}$, the inner structure of scattering materials can be imaged up to $\SI{1}{\milli\metre}$ in depth. A volume can be acquired by repeatedly scanning at neighboring lateral positions. We scan at $32\times 32$ lateral positions. With an A-Scan dimension of $430$ pixels, the raw volumes thus have a size of $32\times 32\times 430$. We downsample the raw OCT volumes to a reduced size of $32\times 32\times 32$ using cubic interpolation due to the high computational and memory requirements of 4D architectures. This leads to similar spatial resolutions in all directions and thus simplified architecture design requirements. Preliminary experiments showed no major improvements with larger spatial resolutions. The volumes cover a field of view (FOV) of $\SI{3}{\milli\metre}\times \SI{3}{\milli\metre}\times \SI{3.5}{\milli\metre}$. Overlaid example volumes are shown in Fig.~\ref{fig:example_vols}. 

\begin{figure}[t]
\begin{center}
   \includegraphics[trim=0 0 0 0.5cm, clip, width=1.0\linewidth]{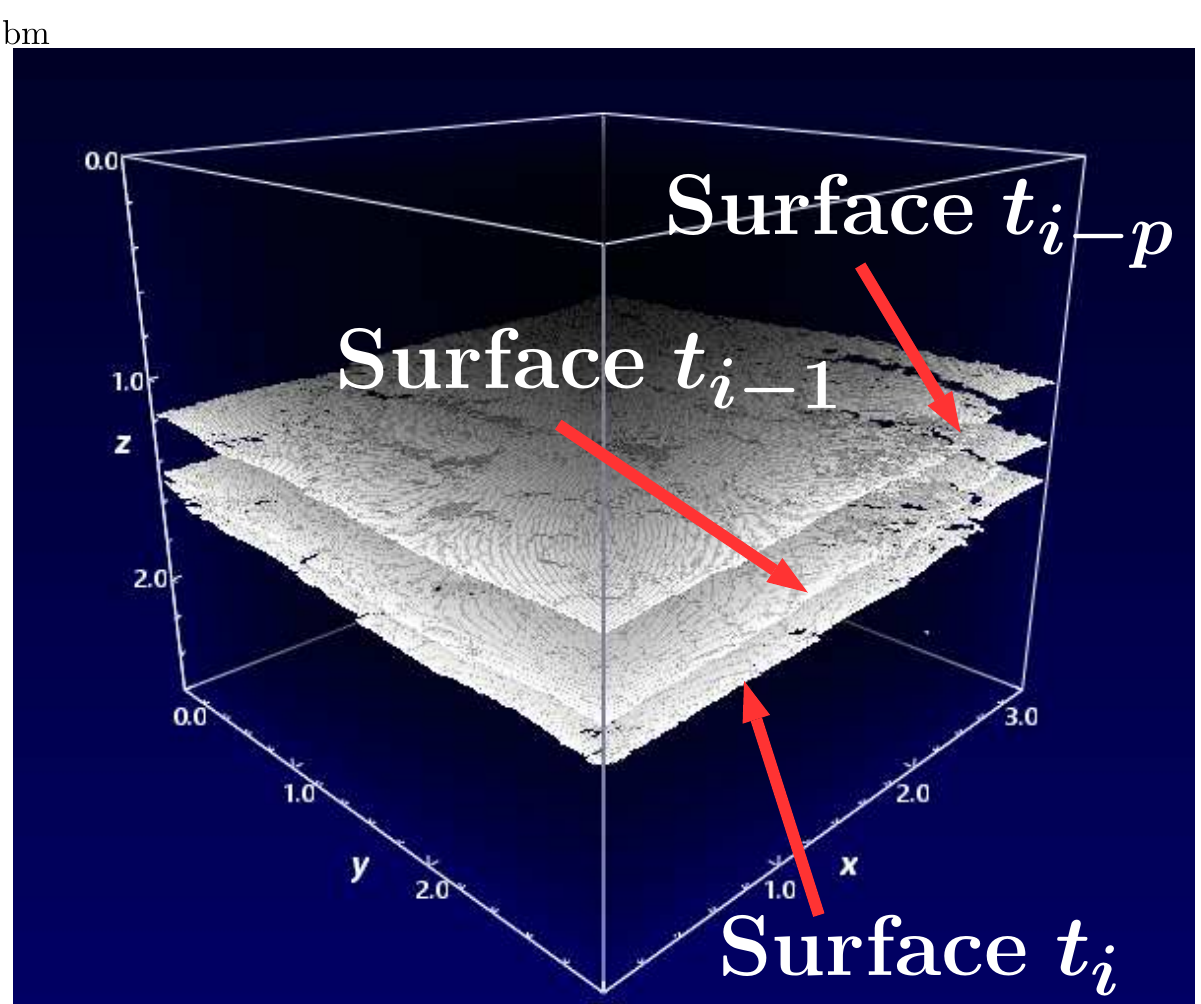}
\end{center}
   \caption{Three pseudo OCT volumes from three time steps.}
\label{fig:example_mips}
\end{figure}

In addition to using streams of the full 3D image volumes, we consider 2D and 3D representations of the tissue surface. Typically, the largest fraction of the infrared light is reflected at the tissue surface. Hence, we employ a maximum intensity projection $P$ to localize the tissue surface in the image volumes, see Figure~\ref{fig:depth_maps}. The surface is either represented by a 2D projection of depth values or by a discrete 3D point cloud, i.e., a volume where voxels representing the surface are set to $1$ and all other voxels are set to $0$. These pseudo volumes are shown in Figure~\ref{fig:example_mips}. To encode the surface as accurately as possible, extraction is performed using the raw OCT volumes with full depth resolution. The resulting depth maps have a size of $32\times 32$. For the pseudo volumes, we reproject the depth maps into volumes of size $32\times 32\times 32$ to match our downsampled volumes in terms of size.

The 2D depth images are a small, efficient representation, however, subsurface information is lost through the projection. They are similar to time-of-flight depth images that have been previously used for force estimation \citep{marban2019recurrent,gao2018learning}. The 3D pseudo volumes encode surface information similar to their 2D counterpart but they are processed in a higher-dimensional space.

\textbf{The phantom} was manufactured using translucent silicone mixed with titan dioxide which leads to light scattering that is similar to the signal found in tissue. 

\textbf{The phantom datasets} were acquired with two types of instrument motion along the needle's shaft. First, we performed a sinusoidal movement with varying amplitudes and frequencies which results in a smooth pattern. For this dataset we acquired $\num{32000}$ samples which we partitioned into sets of $\num{24000}$, $\num{2600}$ and $\num{5400}$ samples for training, validation and testing, respectively (dataset A). Second, we considered movement based on cubic splines. We randomly sampled depth values from $[d_0,d_{max}]$ where $d_0$ is the point of tissue contact and $d_{max}$ the maximum deformation depth along the needle shaft. While being more random, this also provides smooth curves with predictable patterns. Here, we acquired $\num{38000}$ samples which we partition into sets of $\num{31000}$, $\num{2900}$ and $\num{4100}$ samples for training, validation and testing, respectively (dataset B). For both types of acquisition, we acquired all data across multiple experiments with $\approx \num{2500}$ samples each. For each experiment, a new amplitude and frequency (dataset A) or depth pattern (dataset B) was defined. Depths vary from $\SI{1}{\milli\metre}$ to $\SI{3}{\milli\metre}$ and frequencies vary from $\SI{3}{\hertz}$ to $\SI{6}{\hertz}$. We also varied the instrument tip's orientation and relative position with respect to the region of interest (ROI) for each experiment to induce more variation and to avoid overfitting to a particular instrument location or orientation. Similarly, we changed the ROI between experiments to ensure different speckle patterns within the datasets. All data splits are separated by experiments, i.e., all samples of an experiment are only part of one data split. Summarized, datasets A and B represent different types of motion for a homogeneous phantom.

\begin{figure}[t]
\begin{center}
   		   \includegraphics[width=1.0\linewidth]{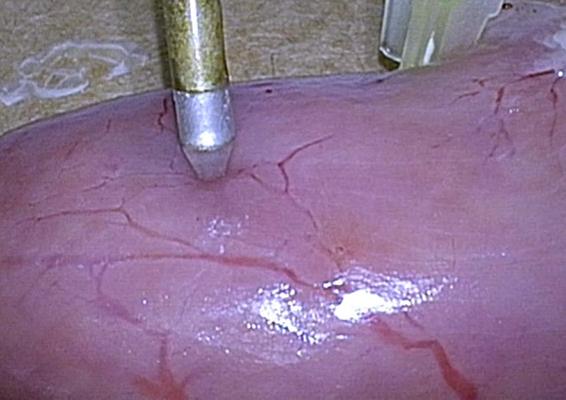}

\end{center}
   \caption{An example image for the tissue being deformed by the needle tool.}
\label{fig:tissue_example}
\end{figure}

\textbf{The tissue dataset} (dataset C) represents a scenario closer to application that we use to test robustness and variation, see Figure~\ref{fig:tissue_example} for an example. In total, we acquired $\num{30000}$ samples for this dataset. We rely on the same movement patterns as for dataset B. Again, we vary the instrument tip's orientation and its relative position with respect to the ROI. Also, we vary the ROI between experiments. In contrast to the phantom, different ROIs can be more different in terms of their subsurface structure relating to fat and muscle composition. Thus, the selection of a single validation/test ROI could induce a bias. Therefore, we perform 20-fold cross-validation (CV) where one experiment corresponding to one ROI is excluded in each experiment. In summary, the tissue dataset is closer to applications and allows us to study the robustness of our approach.

The OCT volumes are acquired at $\SI{60}{\hertz}$ and the force data is acquired at $\SI{500}{\hertz}$. We synchronize both data streams based on timestamps and we assign a force measurement to each OCT volume. We transform the forces from the force sensor's coordinate frame to a coordinate frame located at the needle tip using the known spatial dimensions of the needle and force sensor. Then, we take the magnitude of the resulting force vector as the final label for our learning problem. 

\textbf{Experiment Overview.} All model performances are reported the test set of dataset A and B or the CV result for dataset C. First, we compare our proposed 4D-ST architecture designs and evaluate their performance compared to their 3D-ST counterparts. We relate model performance to architecture efficiency in terms of the number of trainable parameters. Next, we compare different variants of our recurrent models using either LSTMs or GRUs. For better interpretability, we provide a regression plot of the entire force range. Also, we study robustness and variation with tissue dataset C. We calculate processing times by averaging $100$ individual forward passes through a model. Second, we investigate deep learning models using 2D-S, 3D-S, 3D-ST, and 4D-ST data. We consider model versions with different capacity to account for the natural increase in the number of parameters for models with higher-dimensional data processing. Furthermore, we use our ps-4D-ST data derived from depth images to investigate the advantages of 4D-ST data processing. Last, we provide results for different force prediction horizons $f \in \{0,1,2,3,4\}$ and different lengths of temporal history $p \in \{2,4,6,8\}$. As a baseline, all spatio-temporal models use $p=6$. We report the mean absolute error (MAE) in $\si{\milli\newton}$ as an absolute metric and the relative MAE (rMAE) and Pearson's correlation coefficient (PCC) as relative metrics. The rMAE is calculated by dividing the MAE by the ground-truth's standard deviation. For the MAE and rMAE we provide the $25^{\textrm{th}}$ and $75^{\textrm{th}}$ percentile range. We test for significant differences in the median of the absolute errors using the Wilcoxon signed-rank test with $\alpha = 0.05$ significance level.

\section{Results} \label{sec:results}

\begin{table}
\setlength{\tabcolsep}{6pt}
\begin{center}
\begin{tabular}{c l c c c}
 & \textbf{Method} & \textbf{MAE}& \textbf{rMAE} ($\num{e-3}$) & \textbf{PCC} \\
\hline
\parbox[t]{1.5mm}{\multirow{8}{*}{\rotatebox[origin=c]{90}{Dataset A}}} & RN4D & $11.9 (4,16)$ & $42.7 (13,56)$ & $\pmb{0.9984}$ \\
 & facRN4D & $12.3 (4,16)$ & $44.2 (12,56)$ & $0.9982$ \\
 & RN3D-GRU & $19.1 (4,23)$ & $68.4 (16,81)$ & $0.9947$ \\ 
 & cGRU-RN3D & $\pmb{11.7 (3,14)}$ & $\pmb{42.3 (11,52)}$ & $0.9980$ \\
\cline{2-5}
 & RN3D-ST & $21.9 (4,23)$ & $78.8 (15,81)$ & $0.9898$ \\
 & facRN3D & $21.4 (4,21)$ & $76.8 (15,76)$ & $0.9893$ \\
 & RN2D-GRU & $30.8 (7,39)$ & $110 (26,142)$ & $0.9859$ \\ 
 & cGRU-RN2D & $23.3 (5,24)$ & $83.7 (16,85)$ & $0.9885$ \\
\hline \hline
\parbox[t]{1.5mm}{\multirow{8}{*}{\rotatebox[origin=c]{90}{Dataset B}}} & RN4D & $12.0 (4,16)$ & $59.5 (21,81)$ & $0.9968$ \\
 & facRN4D & $13.5 (5,18)$ & $67.1 (24,91)$ & $0.9958$ \\
 & RN3D-GRU & $26.3 (9,35)$ & $130.5 (46,176)$ & $0.9846$ \\ 
 & cGRU-RN3D & $\pmb{10.7 (4,14)}$ & $\pmb{53.2 (18,71)}$ & $\pmb{0.9971}$ \\
\cline{2-5}
 & RN3D-ST & $25.4 (7,31)$ & $125.8 (34,154)$ & $0.9804$ \\
 & facRN3D & $24.8 (7,30)$ & $122.9 (33,147)$ & $0.9809$ \\
 & RN2D-GRU & $35.9 (14,49)$ & $178.3 (68,244)$ & $0.9721$ \\ 
 & cGRU-RN2D & $25.4 (7,32)$ & $126.2 (34,157)$ & $0.9817$ \\
\hline
\end{tabular}
\end{center}
\caption{Comparison of 3D-ST and 4D-ST architectures for both datasets. The MAE is given in $\si{\milli\newton}$. The best values for each dataset are marked bold. RN refers to ResNet and cGRU refers to convGRU. The values in brackets are the $25^{\textrm{th}}$ and $75^{\textrm{th}}$ percentile range.}
\label{tab:4d_res}
\end{table}

\begin{table}
\setlength{\tabcolsep}{6pt}
\begin{center}
\begin{tabular}{c l c c c}
 & \textbf{Method} & \textbf{MAE}& \textbf{rMAE} ($\num{e-3}$) & \textbf{PCC} \\
\hline
\parbox[t]{1.5mm}{\multirow{8}{*}{\rotatebox[origin=c]{90}{Dataset A}}} & RN3D-LSTM & $16.9 (4,19)$ & $60.6 (15,68)$ & $0.9954$ \\
 & cLSTM-RN3D & $11.9 (3,15)$ & $42.9 (12,53)$ & $\pmb{0.9981}$ \\
 & RN3D-GRU & $19.1 (4,23)$ & $68.4 (16,81)$ & $0.9947$ \\ 
 & cGRU-RN3D & $\pmb{11.7 (3,14)}$ & $\pmb{42.3 (11,52)}$ & $0.9980$ \\
\cline{2-5}
 & RN2D-LSTM & $30.5 (6,39)$ & $108.0 (24,139)$ & $0.9862$ \\
 & cLSTM-RN2D & $22.8 (4,24)$ & $81.8 (14,87)$ & $0.9893$ \\
 & RN2D-GRU & $30.8 (7,39)$ & $110.0 (26,142)$ & $0.9859$ \\ 
 & cGRU-RN2D & $23.3 (5,24)$ & $83.7 (16,85)$ & $0.9885$ \\
\hline \hline
\parbox[t]{1.5mm}{\multirow{8}{*}{\rotatebox[origin=c]{90}{Dataset B}}} & RN3D-LSTM & $24.5 (9,32)$ & $121.7 (42,160)$ & $0.9862$ \\
 & cLSTM-RN3D & $10.8 (4,15)$ & $53.6 (18,72)$ & $0.9970$ \\
 & RN3D-GRU & $26.3 (9,35)$ & $130.5 (46,176)$ & $0.9846$ \\ 
 & cGRU-RN3D & $\pmb{10.7 (4,14)}$ & $\pmb{53.2 (18,71)}$ & $\pmb{0.9971}$ \\
\cline{2-5}
 & RN2D-LSTM & $35.5 (13,50)$ & $176.0 (67,249)$ & $0.9733$ \\
 & cLSTM-RN2D & $27.7 (8,36)$ & $137.3 (39,179)$ & $0.9792$ \\
 & RN2D-GRU & $35.9 (14,49)$ & $178.3 (68,244)$ & $0.9721$ \\ 
 & cGRU-RN2D & $25.4 (7,32)$ & $126.2 (34,157)$ & $0.9817$ \\
\hline
\end{tabular}
\end{center}
\caption{Comparison of LSTM-based recurrent models to GRU-based recurrent models. The MAE is given in $\si{\milli\newton}$. The best values for each dataset are marked bold. RN refers to ResNet and cGRU/cLSTM refers to convGRU/convLSTM. The values in brackets are the $25^{\textrm{th}}$ and $75^{\textrm{th}}$ percentile range.}
\label{tab:lstm_vs_gru}
\end{table}

\textbf{4D-ST and 3D-ST Architectures.} The results for all 3D-ST and 4D-ST architectures are shown in Table~\ref{tab:4d_res}. Comparing 4D-ST architectures, across both datasets, ResNet4D and convGRU-ResNet3D perform best. For dataset B, convGRU-ResNet3D performs better as there is a significant difference in the absolute errors. Also, convGRU-ResNet3D performs significantly better than facResNet4D across both datasets. Between all models and datasets, ResNet3D-GRU performs worst. In general, the relative metrics are slightly lower for dataset B with spline-based trajectories. Comparing 3D-ST and 4D-ST architectures, the latter significantly outperform their counterparts across all our proposed architectures.

\begin{figure}
\begin{center}
   \includegraphics[width=1.0\linewidth]{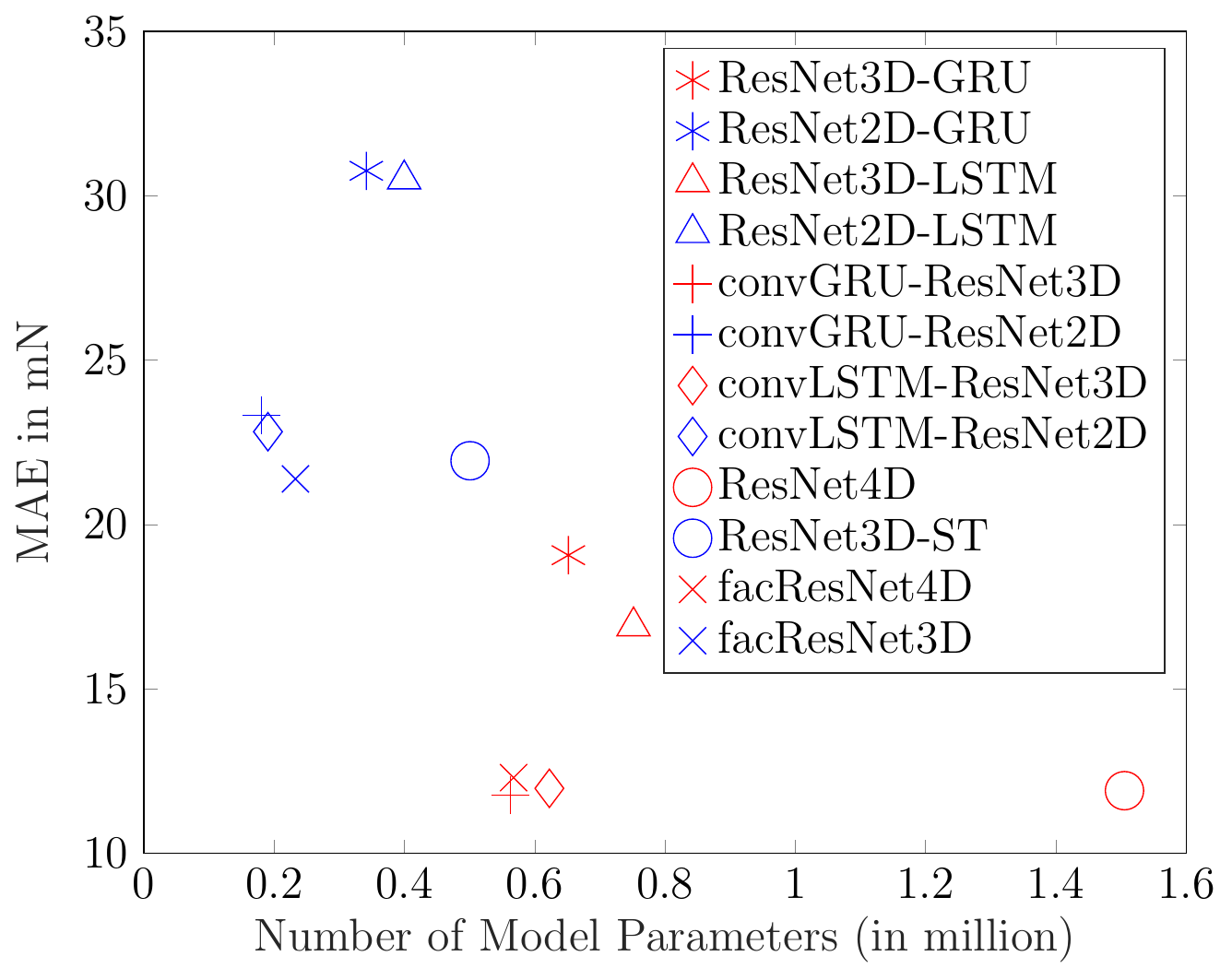}
\end{center}
   \caption{All 4D-ST architectures (red) and their 3D-ST counterpart (blue) in comparison to their MAE and their number of trainable parameters. Models in the lower-left corner have a  lower error and fewer parameters.}
\label{fig:mae_vs_params}
\end{figure}

Next, we show how model capacity relates to performance, see Figure~\ref{fig:mae_vs_params}. While ResNet4D and convGRU-ResNet3D perform similarly, the latter comes with substantially fewer trainable parameters. The 3D-ST models perform significantly worse, however, their capacity in terms of the number of parameters is also lower. 

Furthermore, we consider architecture variants using the more common (convolutional) LSTMs instead of GRUs, see Table~\ref{tab:lstm_vs_gru}. Across 4D and 3D spatio-temporal data, the use of LSTMs leads to similar performance. There is no significant difference between convGRU-ResNet3D and convLSTM-ResNet3D.

\begin{figure}[t]
\begin{center}
   \includegraphics[width=1.0\linewidth]{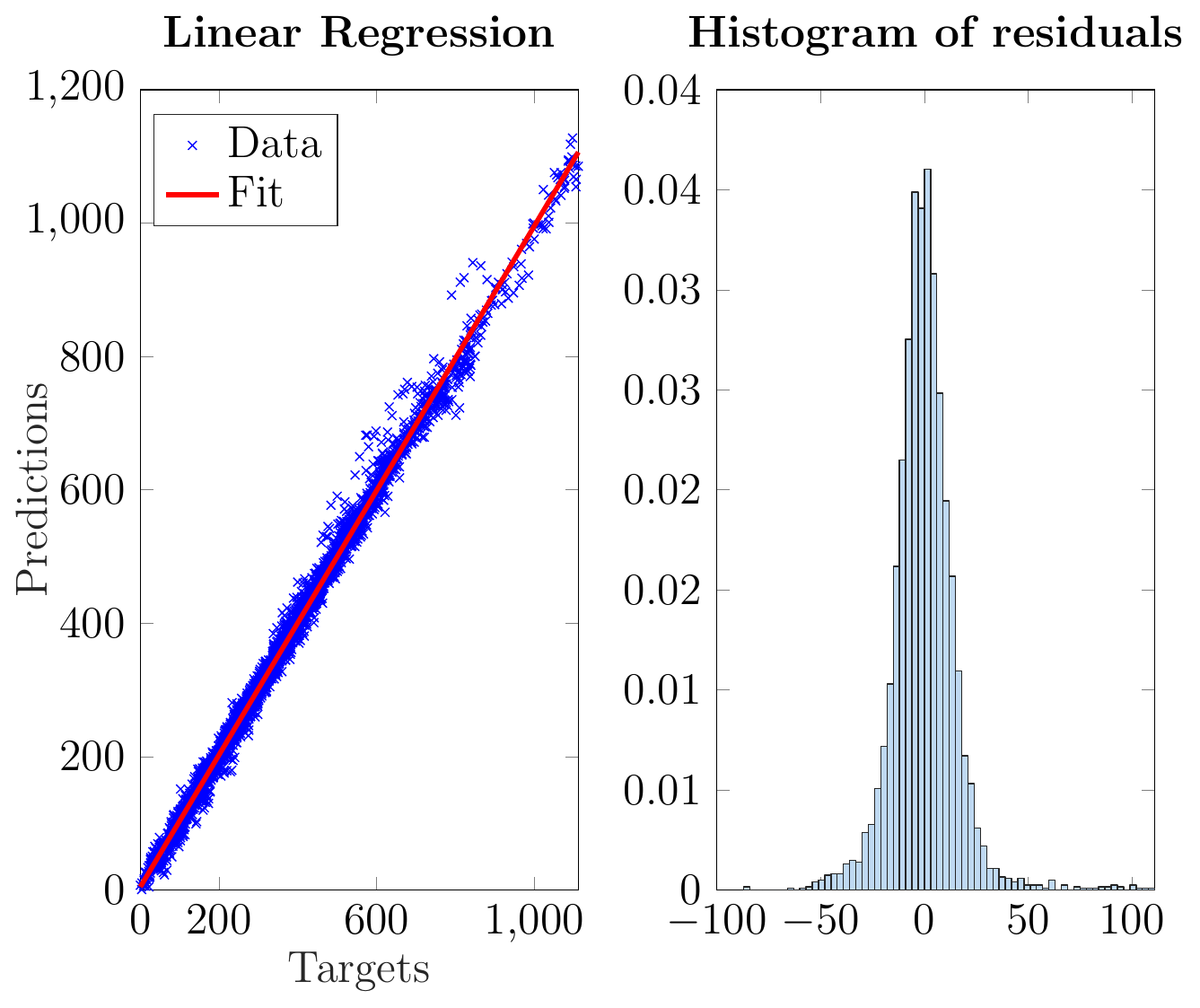}   		 		
\end{center}
\caption{Linear regression plot (left) and the histogram of residuals (right) between targets and predictions, given in $\si{\milli\newton}$. The relationship is significant ($p < 0.05$) with an $R^2$-value of $0.994$. The convGRU-ResNet3D model's predictions and dataset B were used.}
\label{fig:regression}
\end{figure}

For qualitative interpretation, we provide a regression plot between predicted and target force values, see Figure~\ref{fig:regression}. Across the entire force range, predictions closely match the targets. For larger force values closer to $\SI{1}{\newton}$ there are some outliers.

\begin{figure*}[t]
\begin{center}
   \begin{minipage}{0.33\textwidth}
  		\centering
   		\includegraphics[width=1.0\linewidth]{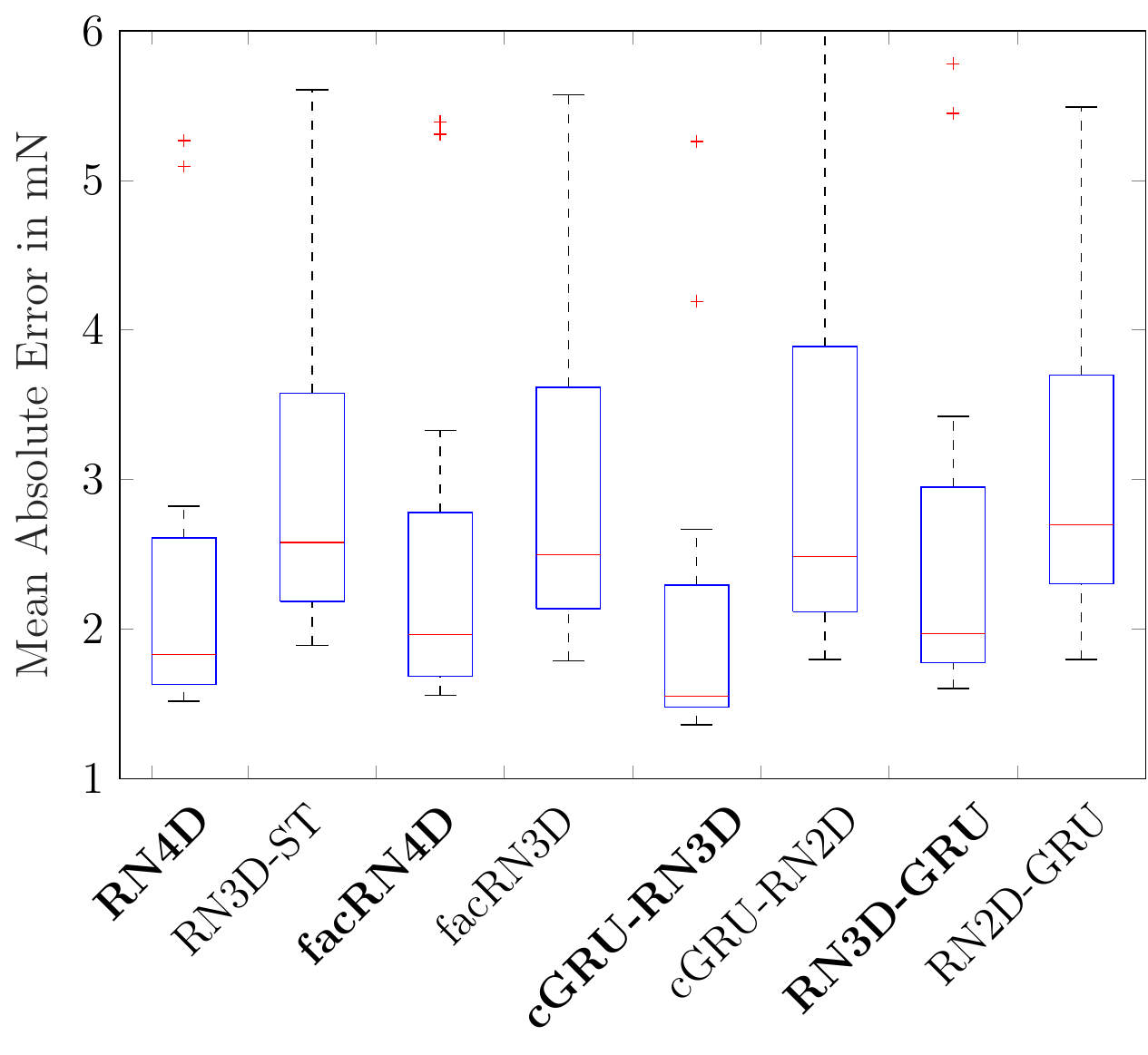}
   \end{minipage}
   \begin{minipage}{0.33\textwidth}
  		\centering   
   		\includegraphics[width=1.0\linewidth]{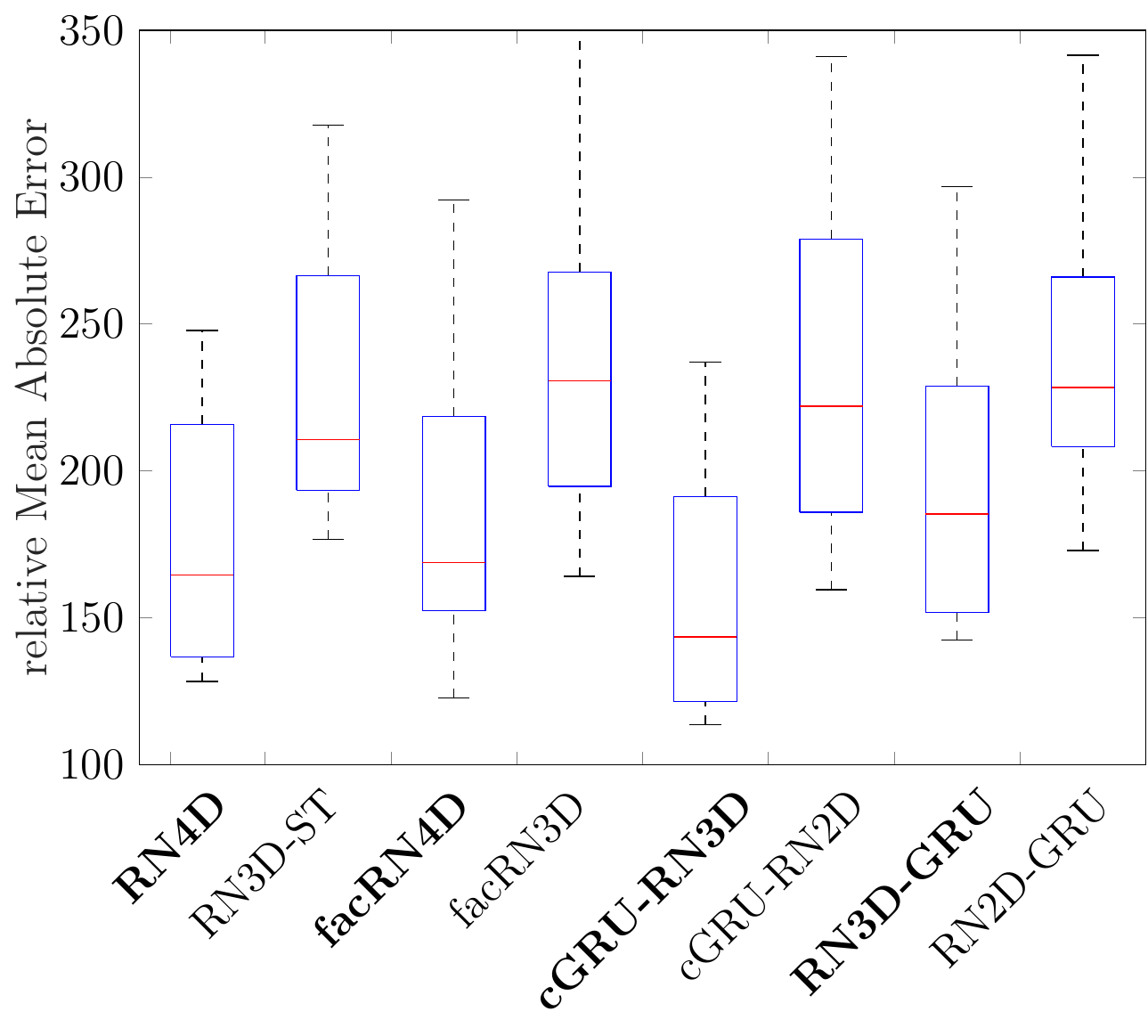}
   \end{minipage}   		
   \begin{minipage}{0.33\textwidth}
  		\centering      
   		\includegraphics[width=1.0\linewidth]{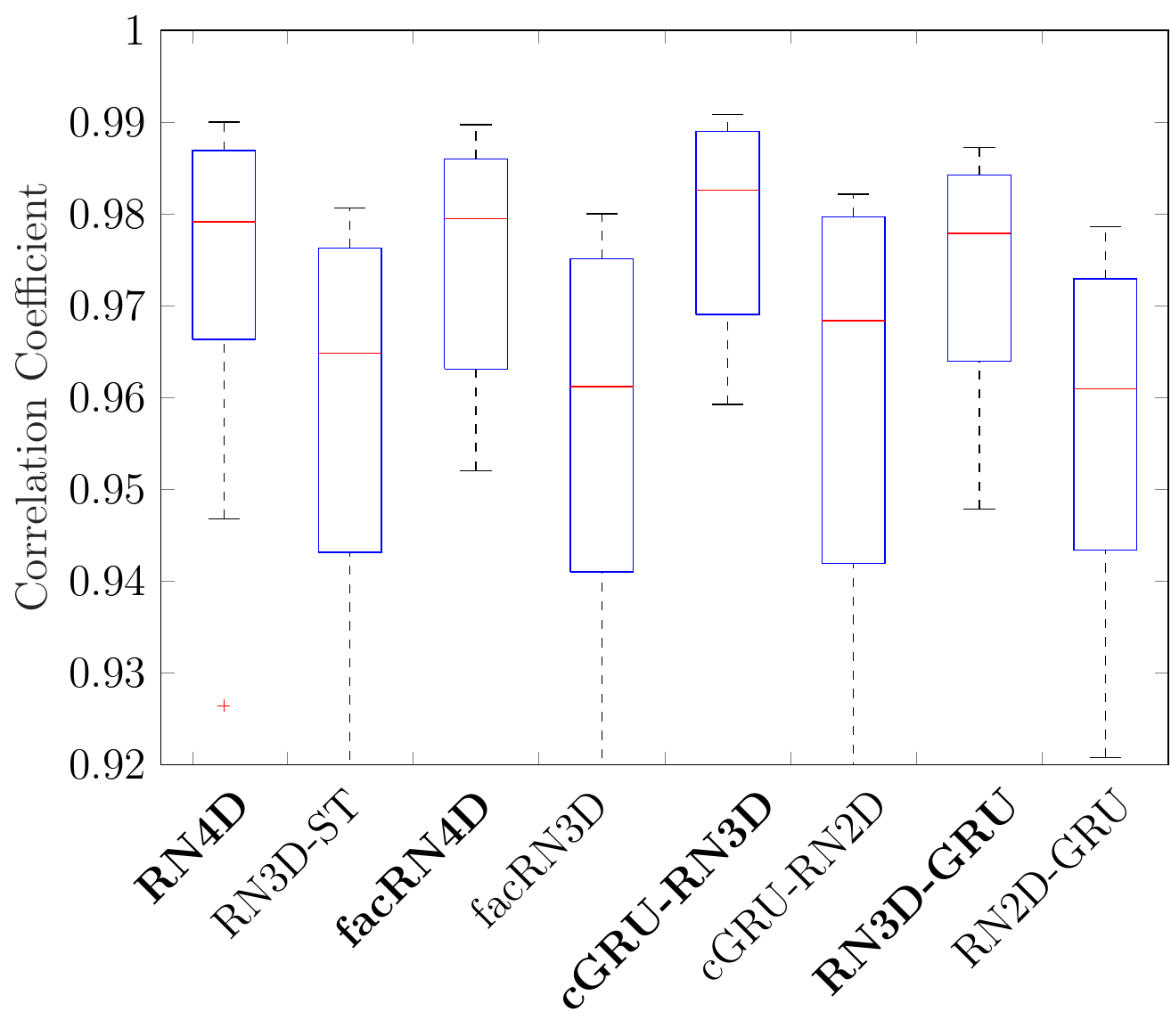}
   \end{minipage}     		 		
\end{center}
   \caption{Boxplots for the three metrics for all 4D-ST (marked bold) and 3D-ST architectures are shown. Each boxplot is generated with $\num{20}$ values from 20-fold CV with our tissue dataset C. RN refers to ResNet and cGRU refers to convGRU.}
\label{fig:tissue}
\end{figure*}

\begin{figure}[t]
\begin{center}
   \includegraphics[width=1.0\linewidth]{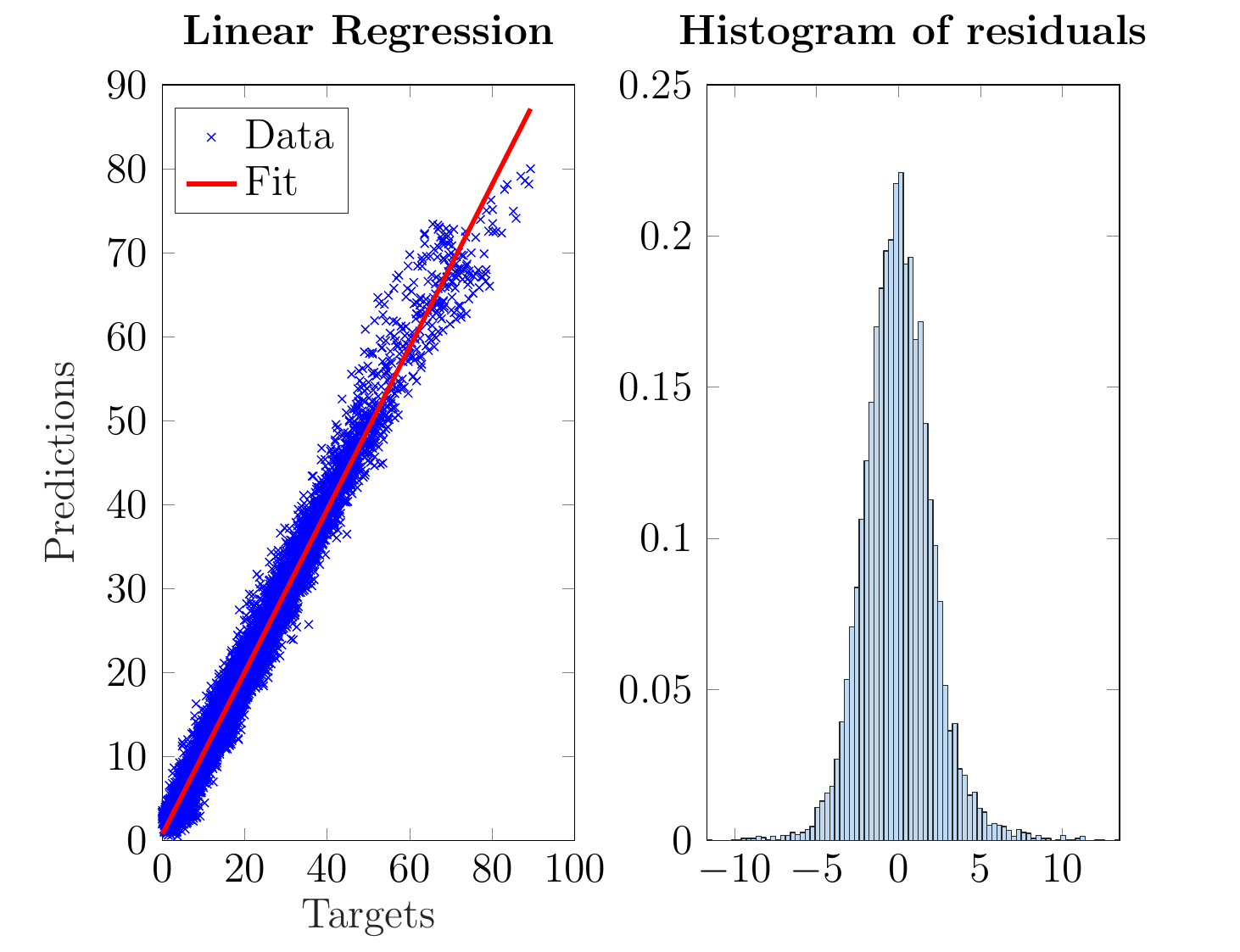}   		 		
\end{center}
\caption{Linear regression plot (left) and the histogram of residuals (right) between targets and predictions, given in $\si{\milli\newton}$. The relationship is significant ($p < 0.05$) with an $R^2$-value of $0.973$. The convGRU-ResNet3D model's predictions and dataset C were used.}
\label{fig:regression_tissue}
\end{figure}

To highlight the advantage of 4D-ST architectures, we also consider a tissue experiment with 20-fold CV to asses variation and robustness, see~\ref{fig:tissue}. In general, the MAE is lower while the relative metrics are worse, compared to the phantom datasets. The size of the boxplots shows that there is some variation across CV folds. Comparing 4D-ST and 3D-ST architectures, the performance difference is similar to the phantom datasets with 4D-ST models performing significantly better. The regression plot in Figure~\ref{fig:regression_tissue} shows the smaller force range for the tissue dataset. Again, predictions match the targets well with a high $R^2$-value.

For real-time applications, processing times are also important. Comparing our best-performing models convGRU-ResNet3D and ResNet4D, both achieve inference times of $\SI{18.2 \pm 2}{\milli\second}$ and $\SI{17.3 \pm 2}{\milli\second}$, respectively. This corresponds to $\SI{55}{\hertz}$ and $\SI{58}{\hertz}$, compared to $\SI{60}{\hertz}$ acquisition speed of the OCT system.

\textbf{Multi-Dimensional Data Representations.} Next, we study how deep learning models with spatial data \citep{otte2016towards,gessert2018force} compare to models using spatio-temporal data. The results are shown in Table~\ref{tab:2d_3d_4d_res}. The 2D-S models perform worst. Adding a temporal dimension (3D-ST) improves performance substantially. The performance increase is statistically significant across all datasets and models. Using volumes (3D-S) for learning performs better than 2D-S or 3D-ST data. Increasing the lower-dimensional models' capacity leads to minor performance improvements. In particular, the ResNet3D-W and ResNet3D-D models come with approximately \num{2000000} and \num{1000000} parameters, respectively. Therefore, they are close to ResNet4D in terms of parameters that comes with approximately \num{1500000} parameters. Overall, the 4D-ST deep learning model ResNet4D performs best, even when compared to 3D models with similar capacity. The performance difference in terms of the median of the absolute errors is statistically significant for all models and datasets.

\begin{table}
\setlength{\tabcolsep}{6.0pt}
\begin{center}
\begin{tabular}{c l c c c}
 & \textbf{Method} & \textbf{MAE}& \textbf{rMAE} ($\num{e-3}$) & \textbf{PCC} \\
\hline
\parbox[t]{1.5mm}{\multirow{10}{*}{\rotatebox[origin=c]{90}{Dataset A}}} & RN2D-S* & $30.7 (7,36)$ & $110 (25,130)$ & $0.9838$ \\
& RN2D-S-W & $15.2 (4,17)$ & $54.7 (13,59)$ & $0.9959$ \\
& RN2D-S-D & $14.9 (4,16)$ & $53.7 (13,58)$ & $0.9959$ \\
\cline{2-5}
& RN3D-S* & $16.2 (4,18)$ & $58 (13,63)$ & $0.9955$ \\
& RN3D-S-W & $15.2 (4,17)$ & $54.7 (13,59)$ & $0.9959$ \\
& RN3D-S-D & $14.9 (4,16)$ & $53.7 (13,58)$ & $0.9959$ \\
\cline{2-5}
& RN3D-ST & $21.9 (4,23)$ & $78.8 (15,81)$ & $0.9898$ \\
& RN3D-ST-W & $21.1 (4,22)$ & $75.6 (14,78)$ & $0.9904$ \\
& RN3D-ST-D & $21.2 (4,22)$ & $76.1 (15,79)$ & $0.9904$ \\
\cline{2-5}
& RN4D & $\pmb{11.9 (4,16)}$ & $\pmb{42.7 (13,56)}$ & $\pmb{0.9984}$ \\
\hline \hline
\parbox[t]{1.5mm}{\multirow{10}{*}{\rotatebox[origin=c]{90}{Dataset B}}} & RN2D-S* & $35.1 (11,48)$ & $174.2 (55,236)$ & $0.9708$ \\
& RN2D-S-W & $33.2 (11,46)$ & $164.6 (53,227)$ & $0.9741$ \\
& RN2D-S-D & $35.3 (11,49)$ & $175.1 (57,242)$ & $0.9704$ \\
\cline{2-5}
& RN3D-S* & $19.7 (6,25)$ & $98 (29,126)$ & $0.9894$ \\
& RN3D-S-W & $18.4 (6,23)$ & $91.3 (27,116)$ & $0.9907$ \\
& RN3D-S-D & $19.2 (6,25)$ & $95.2 (29,123)$ & $0.9900$ \\
\cline{2-5}
& RN3D-ST & $25.4 (7,31)$ & $125.8 (34,154)$ & $0.9804$ \\
& RN3D-ST-W & $23.3 (6,28)$ & $115.7 (32,140)$ & $0.9832$ \\
& RN3D-ST-D & $23.1 (6,27)$ & $114.6 (29,136)$ & $0.9825$ \\
\cline{2-5}
& RN4D & $\pmb{12.0 (4,16)}$ & $\pmb{59.5 (21,80)}$ & $\pmb{0.9968}$ \\
\hline
\end{tabular}
\end{center}
\caption{Comparison of models using 2D-S, 3D-S, 3D-ST and 4D-ST data. \textit{-W} denotes models with more feature maps per layer and \textit{-D} describes models with more layers. The MAE is given in $\si{\milli\newton}$. Models marked by a * were proposed in \citep{gessert2018force}. RN refers to ResNet. The values in brackets are the $25^{\textrm{th}}$ and $75^{\textrm{th}}$ percentile range.}
\label{tab:2d_3d_4d_res}
\end{table}

Moreover, we consider pseudo 4D-ST as a higher-dimensional encoding of the depth images. The results are shown in Table~\ref{tab:pseudo_4d}. Overall, the pseudo 4D-ST models' performance is much closer to the 3D-ST models than to the 4D-ST models. However, the performance difference is statistically significant across both datasets for ResNet4D, facResNet4D, and convGRU-ResNet3D. For ResNet3D-GRU the performance deteriorates. 

\begin{table}
\setlength{\tabcolsep}{3.50pt}
\begin{center}
\begin{tabular}{c l c c c}
 & \textbf{Method} & \textbf{MAE}& \textbf{rMAE} ($\num{e-3}$) & \textbf{PCC} \\
\hline
\parbox[t]{1.5mm}{\multirow{8}{*}{\rotatebox[origin=c]{90}{Dataset A}}} & ps-RN4D & $18.7 (4,21)$ & $67.1 (15,76)$ & $0.9942$ \\
 & ps-facRN4D & $17.9 (4,21)$ & $64.1 (15,75)$ & $0.9943$ \\
 & ps-RN3D-GRU & $43.3 (9,55)$ & $155 (33,196)$ & $0.9848$ \\
 & ps-cGRU-RN3D & $\pmb{15.0 (4,17)}$ & $\pmb{53.8 (13,61)}$ & $\pmb{0.9959}$ \\
\cline{2-5}
 & RND-ST & $21.9 (4,23)$ & $78.8 (15,81)$ & $0.9898$ \\
 & facRN3D & $21.4 (4,21)$ & $76.8 (15,76)$ & $0.9893$ \\
 & RN2D-GRU & $30.8 (7,39)$ & $110 (26,142)$ & $0.9859$ \\ 
 & cGRU-RN2D & $23.3 (5,24)$ & $83.7 (16,85)$ & $0.9885$ \\
\hline \hline
\parbox[t]{1.5mm}{\multirow{8}{*}{\rotatebox[origin=c]{90}{Dataset B}}} & ps-RN4D & $17.8 (6,25)$ & $88.2 (31,122)$ & $0.9930$ \\
 & ps-facRN4D & $19.2 (7,26)$ & $95.2 (33,130)$ & $0.9915$ \\
 & ps-RN3D-GRU & $37.8 (15,50)$ & $188 (73,247)$ & $0.9705$ \\ 
 & ps-cGRU-RN3D & $\pmb{15.4 (5,21)}$ & $\pmb{76.5 (25,102)}$ & $\pmb{0.9940}$ \\
\cline{2-5}
 & RN3D-ST & $25.4 (7,31)$ & $126 (34,154)$ & $0.9804$ \\
 & facRN3D & $24.8 (7,30)$ & $123 (33,147)$ & $0.9809$ \\
 & RN2D-GRU & $35.9 (14,49)$ & $178 (68,244)$ & $0.9721$ \\ 
 & cGRU-RN2D & $25.4 (7,32)$ & $126 (34,157)$ & $0.9817$ \\
\hline
\end{tabular}
\end{center}
\caption{Comparison of 3D-ST and pseudo 4D-ST data representations for both datasets. Pseudo 4D-ST models are labeled by the prefix ps. The MAE is given in $\si{\milli\newton}$. The best values for each dataset are marked bold. RN refers to ResNet. The values in brackets are the $25^{\textrm{th}}$ and $75^{\textrm{th}}$ percentile range.}
\label{tab:pseudo_4d}
\end{table}

\begin{figure*}[t]
\begin{center}
   \begin{minipage}{0.4\textwidth}
  		\centering
   		\includegraphics[width=0.8\linewidth]{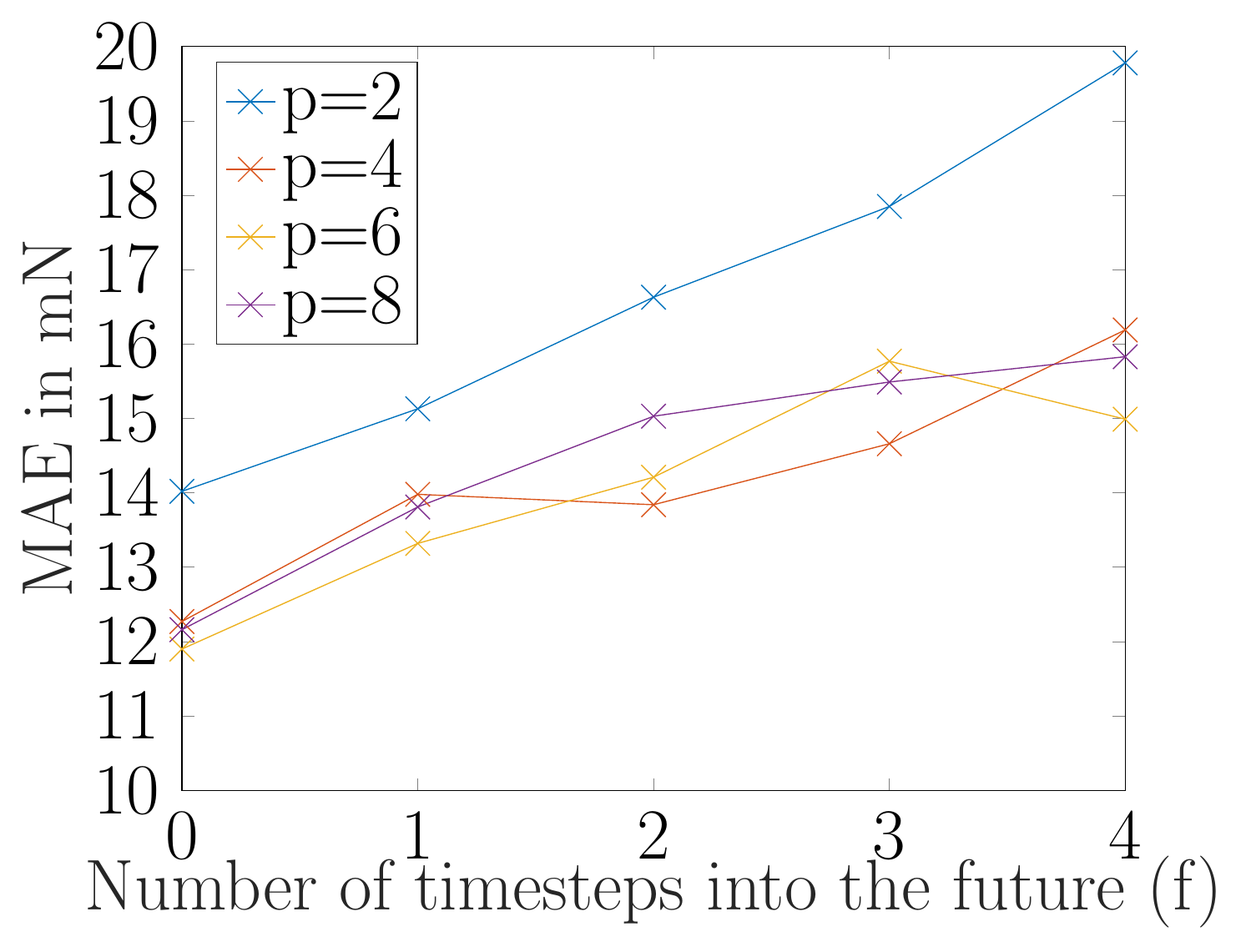}
    	\subcaption[2nd caption]{ResNet4D A}
   \end{minipage}
   \begin{minipage}{0.4\textwidth}
  		\centering   
   		\includegraphics[width=0.8\linewidth]{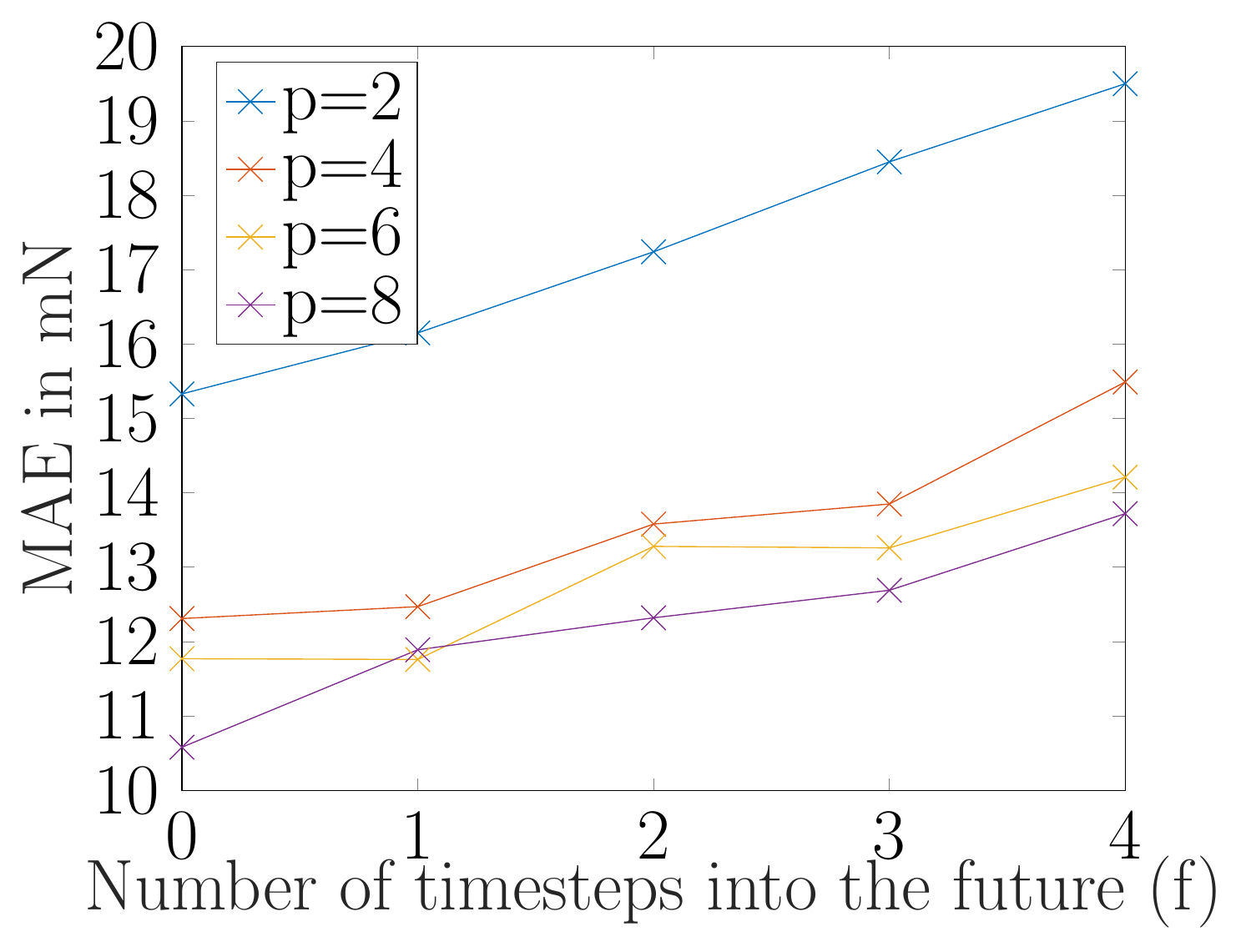}
    	\subcaption[2nd caption]{convGRU-ResNet3D A}
   \end{minipage}   		
   \begin{minipage}{0.4\textwidth}
  		\centering      
   		\includegraphics[width=0.8\linewidth]{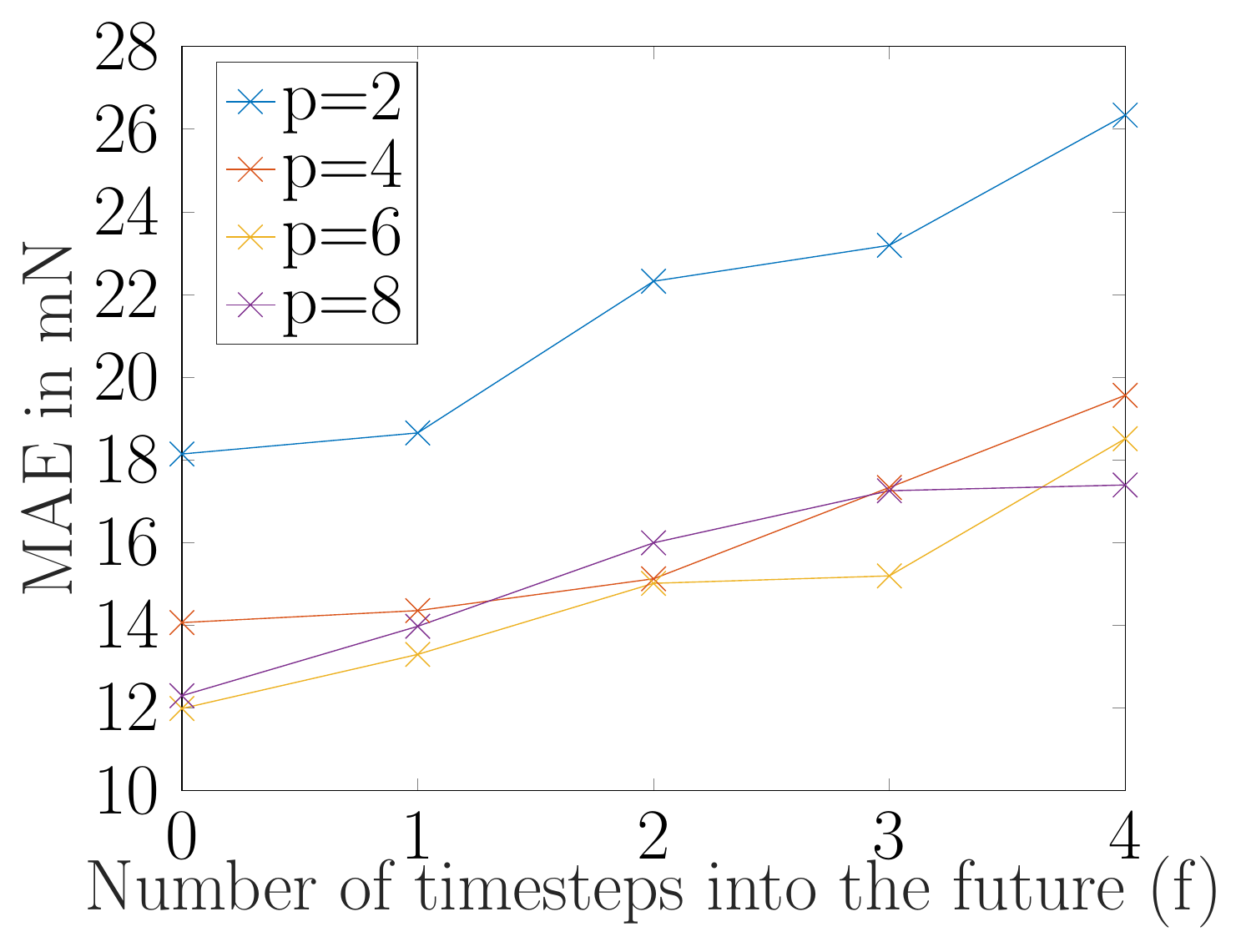}
    	\subcaption[2nd caption]{ResNet4D B}
   \end{minipage}     		
   \begin{minipage}{0.4\textwidth}
  		\centering     
   		\includegraphics[width=0.8\linewidth]{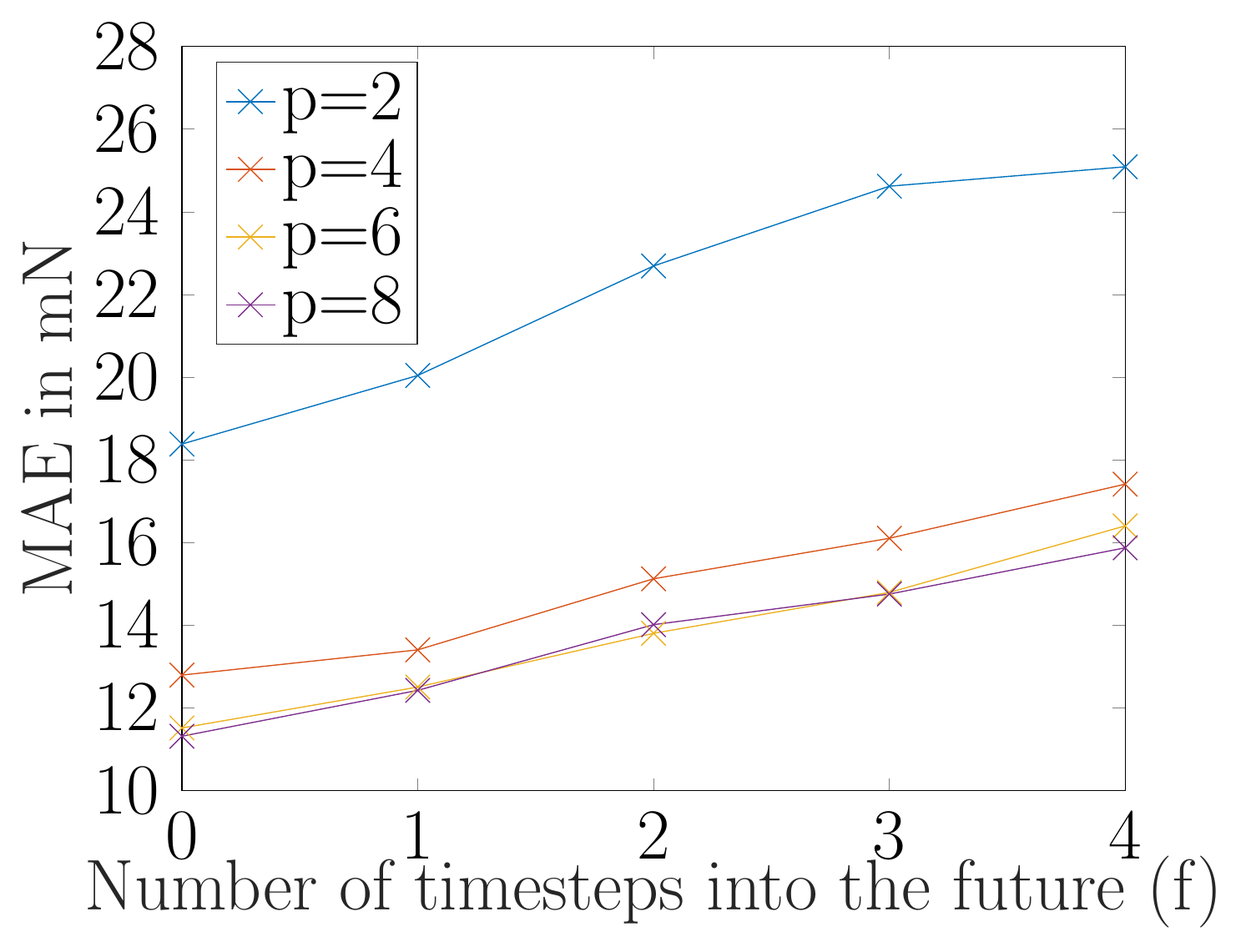}
    	\subcaption[2nd caption]{convGRU-ResNet3D B}
   \end{minipage}    		
\end{center}
   \caption{Force estimation and prediction results for varying $p$ and $f$ with two datasets A and B with architectures ResNet4D and convGRU-ResNet3D. For each setting, a new model is trained and evaluated.}
\label{fig:prediction}
\end{figure*}

\textbf{Temporal Information and Force Prediction.} Last, we investigate the temporal properties of the two top-performing models ResNet4D and convGRU-ResNet3D. The results for variations of the temporal history $p$ and prediction horizon $f$ are shown in Figure~\ref{fig:prediction}. For each combination of $p$ and $f$ a model was trained. Considering force estimation ($f=0$), adding more temporal history improves performance. The largest improvement can be observed between $p=2$ and $p=4$. Then, performance tends to saturate. Similar observations can be made for $f>0$ and increasing values for $p$. When increasing $f$, prediction works well as the MAE only increases by $\approx \SI{25}{\percent}$ (DS A) and $\approx \SI{39}{\percent}$ (DS B) between $f=0$ and $f=4$ for $p \in \{4,6,8\}$ with convGRU-ResNet3D. Comparing ResNet4D and convGRU-ResNet3D, both show similar trends for increasing $f$. However, for increasing $p$, convGRU-ResNet3D shows a consistent performance improvement for all $f$ and both datasets, while ResNet4D shows varying results for $p \in \{4,6,8\}$ across all $f$.

\begin{figure}
\begin{center}
   \includegraphics[width=1.0\linewidth]{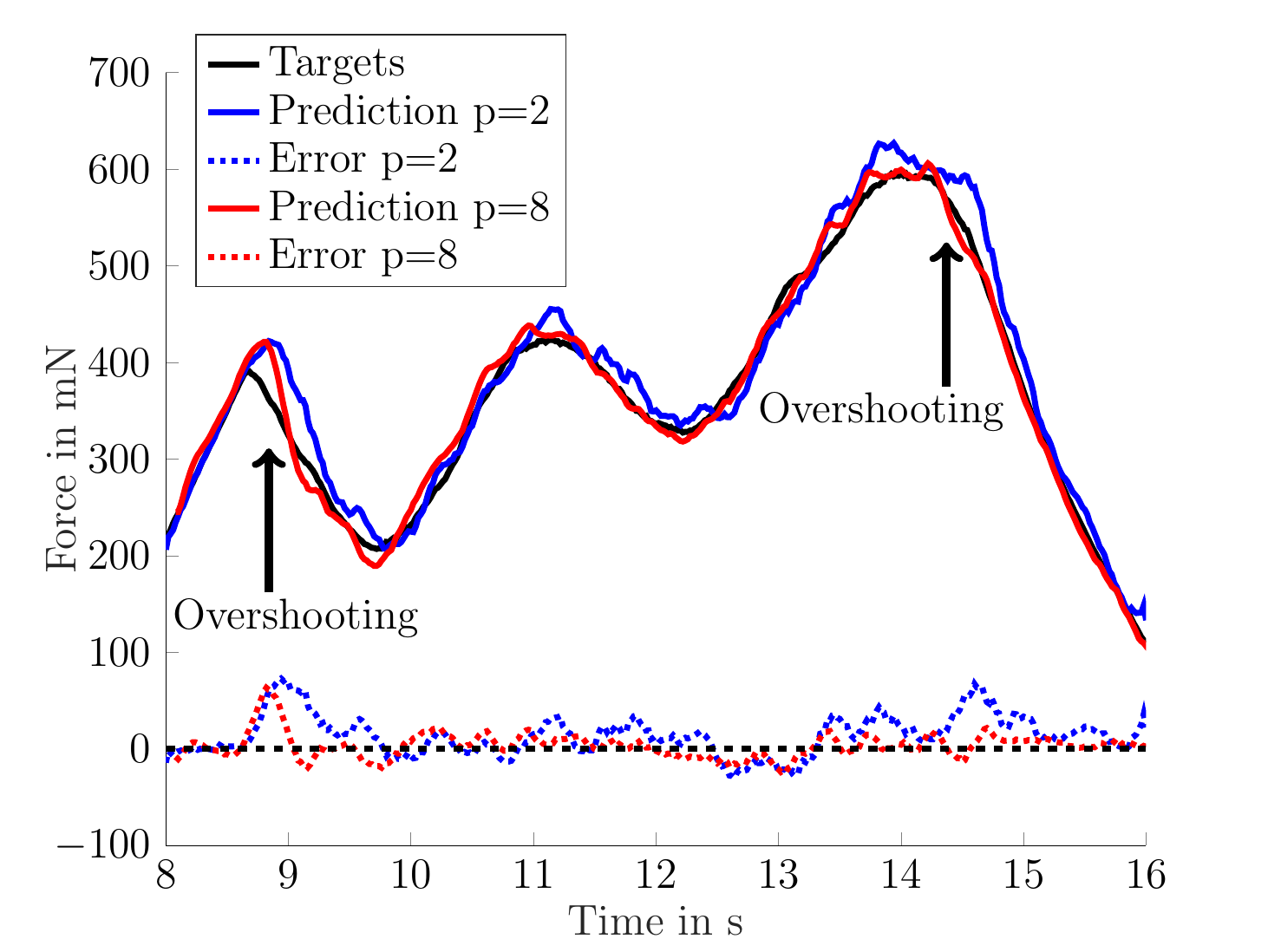}
\end{center}
   \caption{Example for the force trend when predicting four time steps into the future using convGRU-ResNet3D and a history of $p=2$ or $p=8$ time steps. At sudden force trend changes, we can observe overshooting of the predictions.}
\label{fig:qual}
\end{figure}

Last, we also provide qualitative results for predicting forces four time steps into the future with different lengths of history, see Figure~\ref{fig:qual}. For a longer history, predictions are more consistent. For sudden changes in the force trend, overshooting can be observed.

\section{Discussion} \label{sec:discussion}

In this work, we introduce 4D deep learning with 4D spatio-temporal data for OCT-based force estimation. As 4D deep learning has not been studied for either OCT or force estimation, we designed and evaluated four different architectures with different concepts for 4D spatio-temporal data processing. Overall, our convGRU-ResNet3D which performs temporal processing first, followed by spatial processing, shows the best performance. This is particularly interesting as a lot of spatio-temporal deep learning approaches in the natural image domain perform spatial feature extraction first, followed by temporal processing of the spatial features (ResNet3D-GRU) \citep{asadi2017survey}. However, other force estimation methods on 2D data have shown similar observations as ours \citep{gessert2018needle,gessert2019spatio}. In addition, ResNet4D shows performance similar to convGRU-ResNet3D, however, the model comes with substantially more parameters. Decomposing the convolutions into a spatial and a temporal part with facResNet4D improves this aspect but reduces performance. Also, when replacing GRUs with traditional LSTMs, performance remains similar while the number of model parameters increases slightly. Thus, overall, convGRU-ResNet3D represents a high-performing and efficient deep learning architecture for 4D OCT-based force estimation.

To highlight the value of 4D spatio-temporal data processing, we compared our approaches to their 3D spatio-temporal counterparts. These models use depth images extracted from volumes which is similar to previous force estimation approaches relying on depth representations from time-of-flight cameras \citep{gao2018learning}. We find that across all our concepts for spatio-temporal processing, the 4D models consistently outperform their respective 3D counterpart with a statistically significant performance difference both for phantom and tissue data. This suggests that using 4D data instead of 3D data is beneficial for our problem. Notably, although the 4D models require increased processing times, our best performing models convGRU-ResNet3D and ResNet4D still achieve inference times close to our OCT system's acquisition rate. This suggests that real-time applications are also feasible with 4D deep learning models.

While other 4D spatio-temporal CNN applications with different medical imaging modalities \citep{clark2019convolutional,myronenko20194d} did not show improvements over lower-dimensional approaches, we highlight the value of utilizing full 4D information when it is available. For OCT data this is particularly interesting since devices for 4D OCT imaging with high temporal resolution are available \citep{siddiqui2018high} but there have been no studies utilizing the full 4D information in deep learning models. Thus our insights could improve other medical OCT applications where 4D OCT data is available.

Considering absolute and relative metric results, absolute MAE values are slightly larger for phantom experiments compared to tissue experiments due to a larger force range as shown in Figure~\ref{fig:regression}. While this force range is typical for applications such as lung tumor localization \citep{mccreery2008feasibility}, other applications such as retinal microsurgery require a smaller force range and higher resolution \citep{gupta1999surgical}. Our tissue experiments demonstrate that our approach is also scalable to smaller force ranges, see Figure~\ref{fig:regression_tissue}. Due to the larger variability between tissue samples compared to phantom data, estimation appears to be more difficult as relative metrics are slightly lower than for the phantom experiments. Nevertheless, regarding absolute values, the MAE is around $\SI{2}{\milli\newton}$ for our 4D-ST architectures. This suggests the suitability of our approach for a variety of clinical applications where force sensing can be helpful and different force ranges are required \citep{trejos2010force}.

Given the same high-level architecture, deep learning models that process lower-dimensional data contain fewer parameters, thus, having a lower representational power. Therefore, we also evaluate the four CNN-based lower-dimensional data processing models which have been employed previously \citep{gessert2018force,marban2019recurrent} with increased model capacity. Even when adding more layers to the models or increasing the number of feature maps, the ResNet3D-D and ResNet3D-W model variants show a significantly lower performance than ResNet4D. This demonstrates that there is an inherent advantage of using 4D data which is not connected to the natural increase in model capacity for higher-dimensional data processing methods.

The high performance of models using 4D-ST data leads to the question of whether the advantage is caused by richer subsurface information being present or processing in a higher-dimensional space. The depth maps we use have a smaller size than the volumes which makes an immediate comparison to larger volumes difficult. Therefore, we disentangle the aspects of input size and subsurface information by using pseudo volumes that encode the depth images in a volume that has the same size as our full OCT volumes. These pseudo 3D volumes encode the same information as the 2D depth images. Overall, we find that pseudo 4D-ST data processing performance is closer to 3D-ST data than 4D-ST data which indicates that the majority of the advantage can be traced back to richer information. However, using pseudo 4D-ST data leads to a statistically significant performance improvement for three of our four architectures across both datasets. This indicates that there is an inherent advantage of using higher-dimensional data representations for our problem. Intuitively, this relates to the well-known kernel trick where the same fundamental principle is exploited. Also, in the natural image domain, point clouds have been encoded as volumes for deep learning applications \citep{maturana2015voxnet}. This insight comes with important implications for other force estimation approaches. 4D data processing is not only advantageous when native 4D data is available, but performance can also be improved by transforming streams of 2D depth representations into a higher-dimensional space. Thus, previous force estimation approaches using RGBD-based 3D-ST depth data could potentially benefit from our insights on pseudo 4D-ST data.

Last, we also provide a detailed analysis of the temporal dimension, see Figure~\ref{fig:prediction}. In terms of quantitative results, we find that an increasing length of the history $p$ leads to improvements for ResNet4D and convGRU-ResNet3D both for estimation and prediction which is particularly consistent for the latter. This indicates that the model is very effective at utilizing temporal information.

Previous approaches for force estimation have used temporal information successfully as well \citep{aviles2017towards,gao2018learning}, however, they have not attempted force prediction. Given a smooth change in forces and deformation, forces should be predictable which we demonstrate successfully for different prediction horizons. Due to the volatile nature of movement during surgery, long-term prediction is not reasonable, however, for our time horizon of $\SI{16}{\milli\second}$ to $\SI{64}{\milli\second}$ prediction should be possible. Still there are limitations to predictability, in particular, for unexpected changes. We demonstrate this aspect in Figure~\ref{fig:qual} where sudden changes in the force trend visibly impact prediction. At the same time, a longer history appears to help in obtaining more consistent estimates. However, a very long history could also have a downside if old, irrelevant time steps dominate the current force prediction. Thus, future work could also consider weighting techniques to put more emphasis on parts of the history that are relevant for force prediction.

For clinical application, force prediction could be useful in the context of fully automated robotic interventions \citep{haidegger2009force} as force prediction would be useful for safety features. An automatic system could stop before certain force thresholds with a risk of trauma are exceeded \citep{haouchine2018vision}. 

Overall, we provide an in-depth analysis of multi-dimensional deep learning for OCT-based force estimation. We present important insights on image data dimensionality, design and evaluate several 4D deep learning architectures and find multiple significant performance improvements over previous methods. Our insight could benefit both medical applications where 4D data is available and force estimation approaches that rely on lower-dimensional data so far. 

\section{Conclusion} \label{sec:conclusion}

In this work, we propose 4D spatio-temporal deep learning for OCT-based force estimation. In this context, we provide a comprehensive study on network architectures and data representations. In particular, we design and evaluate four architectures with different principles for spatial and temporal processing. We find that decoupling spatial and temporal processing with a convGRU-ResNet3D performs well across multiple scenarios. Results of our study regarding different image data representations indicate that 4D deep learning using the full spatial and temporal information is preferable over 3D data. Furthermore, three-dimensional representations of surface data resulted in better performance compared to two-dimensional representations, which may be interesting for other surface-based force estimation approaches. Finally, we find that exploiting the temporal dimension is advantageous for estimation and that short-term force prediction is also feasible. Future work could extend our approach to predicting a full force vector. Also, our proposed methods could be applied to other biomedical applications where 4D image data is available.

\section*{Acknowledgment}

This work was partially funded by the TUHH $i^3$ initiative.

\bibliographystyle{model2-names.bst}\biboptions{authoryear}
\bibliography{egbib}

\begin{thebibliography}{58}
\expandafter\ifx\csname natexlab\endcsname\relax\def\natexlab#1{#1}\fi
\providecommand{\url}[1]{\texttt{#1}}
\providecommand{\href}[2]{#2}
\providecommand{\path}[1]{#1}
\providecommand{\DOIprefix}{doi:}
\providecommand{\ArXivprefix}{arXiv:}
\providecommand{\URLprefix}{URL: }
\providecommand{\Pubmedprefix}{pmid:}
\providecommand{\doi}[1]{\href{http://dx.doi.org/#1}{\path{#1}}}
\providecommand{\Pubmed}[1]{\href{pmid:#1}{\path{#1}}}
\providecommand{\bibinfo}[2]{#2}
\ifx\xfnm\relax \def\xfnm[#1]{\unskip,\space#1}\fi
\bibitem[{Abadi et~al.(2015)Abadi, Agarwal, Barham, Brevdo, Chen and
  et~al.}]{tensorflow2015-whitepaper}
\bibinfo{author}{Abadi, M.}, \bibinfo{author}{Agarwal, A.},
  \bibinfo{author}{Barham, P.}, \bibinfo{author}{Brevdo, E.},
  \bibinfo{author}{Chen, Z.}, \bibinfo{author}{et~al., C.C.},
  \bibinfo{year}{2015}.
\newblock \bibinfo{title}{{TensorFlow}: Large-scale machine learning on
  heterogeneous systems}.
\newblock \URLprefix \url{http://tensorflow.org/}. \bibinfo{note}{software
  available from tensorflow.org}.
\bibitem[{Asadi-Aghbolaghi et~al.(2017)Asadi-Aghbolaghi, Clapes, Bellantonio,
  Escalante, Ponce-L{\'o}pez and Bar{\'o}~et al.}]{asadi2017survey}
\bibinfo{author}{Asadi-Aghbolaghi, M.}, \bibinfo{author}{Clapes, A.},
  \bibinfo{author}{Bellantonio, M.}, \bibinfo{author}{Escalante, H.J.},
  \bibinfo{author}{Ponce-L{\'o}pez, V.}, \bibinfo{author}{Bar{\'o}~et al., X.},
  \bibinfo{year}{2017}.
\newblock \bibinfo{title}{A survey on deep learning based approaches for action
  and gesture recognition in image sequences}, in:
  \bibinfo{booktitle}{International Conference on Automatic Face \& Gesture
  Recognition}, \bibinfo{organization}{IEEE}. pp. \bibinfo{pages}{476--483}.
\bibitem[{Aviles et~al.(2017)Aviles, Alsaleh, Hahn and
  Casals}]{aviles2017towards}
\bibinfo{author}{Aviles, A.I.}, \bibinfo{author}{Alsaleh, S.M.},
  \bibinfo{author}{Hahn, J.K.}, \bibinfo{author}{Casals, A.},
  \bibinfo{year}{2017}.
\newblock \bibinfo{title}{Towards retrieving force feedback in robotic-assisted
  surgery: A supervised neuro-recurrent-vision approach}.
\newblock \bibinfo{journal}{IEEE Transactions on Haptics} \bibinfo{volume}{10},
  \bibinfo{pages}{431--443}.
\bibitem[{Aviles et~al.(2015)Aviles, Alsaleh, Sobrevilla and
  Casals}]{aviles2015force}
\bibinfo{author}{Aviles, A.I.}, \bibinfo{author}{Alsaleh, S.M.},
  \bibinfo{author}{Sobrevilla, P.}, \bibinfo{author}{Casals, A.},
  \bibinfo{year}{2015}.
\newblock \bibinfo{title}{Force-feedback sensory substitution using supervised
  recurrent learning for robotic-assisted surgery}, in:
  \bibinfo{booktitle}{International Conference of the IEEE Engineering in
  Medicine and Biology Society}, \bibinfo{organization}{IEEE}. pp.
  \bibinfo{pages}{1--4}.
\bibitem[{Bengs et~al.(2019)Bengs, Gessert and Schlaefer}]{bengs2019}
\bibinfo{author}{Bengs, M.}, \bibinfo{author}{Gessert, N.},
  \bibinfo{author}{Schlaefer, A.}, \bibinfo{year}{2019}.
\newblock \bibinfo{title}{4d spatio-temporal deep learning with 4d fmri data
  for autism spectrum disorder classification}, in:
  \bibinfo{booktitle}{International Conference on Medical Imaging with Deep
  Learning}.
\bibitem[{Cho et~al.(2014)Cho, Van~Merri{\"e}nboer, Gulcehre, Bahdanau,
  Bougares, Schwenk and Bengio}]{cho2014learning}
\bibinfo{author}{Cho, K.}, \bibinfo{author}{Van~Merri{\"e}nboer, B.},
  \bibinfo{author}{Gulcehre, C.}, \bibinfo{author}{Bahdanau, D.},
  \bibinfo{author}{Bougares, F.}, \bibinfo{author}{Schwenk, H.},
  \bibinfo{author}{Bengio, Y.}, \bibinfo{year}{2014}.
\newblock \bibinfo{title}{Learning phrase representations using rnn
  encoder-decoder for statistical machine translation}.
\newblock \bibinfo{journal}{arXiv preprint arXiv:1406.1078} .
\bibitem[{Choy et~al.(2019)Choy, Gwak and Savarese}]{choy20194d}
\bibinfo{author}{Choy, C.}, \bibinfo{author}{Gwak, J.},
  \bibinfo{author}{Savarese, S.}, \bibinfo{year}{2019}.
\newblock \bibinfo{title}{4d spatio-temporal convnets: Minkowski convolutional
  neural networks}, in: \bibinfo{booktitle}{{Conference on Computer Vision and
  Pattern Recognition}}, pp. \bibinfo{pages}{3075--3084}.
\bibitem[{Clark and Badea(2019)}]{clark2019convolutional}
\bibinfo{author}{Clark, D.}, \bibinfo{author}{Badea, C.}, \bibinfo{year}{2019}.
\newblock \bibinfo{title}{Convolutional regularization methods for 4d, x-ray ct
  reconstruction}, in: \bibinfo{booktitle}{Medical Imaging 2019: Physics of
  Medical Imaging}, \bibinfo{organization}{International Society for Optics and
  Photonics}. p. \bibinfo{pages}{109482A}.
\bibitem[{Cooijmans et~al.(2016)Cooijmans, Ballas, Laurent, G{\"u}l{\c{c}}ehre
  and Courville}]{cooijmans2016recurrent}
\bibinfo{author}{Cooijmans, T.}, \bibinfo{author}{Ballas, N.},
  \bibinfo{author}{Laurent, C.}, \bibinfo{author}{G{\"u}l{\c{c}}ehre,
  {\c{C}}.}, \bibinfo{author}{Courville, A.}, \bibinfo{year}{2016}.
\newblock \bibinfo{title}{Recurrent batch normalization}.
\newblock \bibinfo{journal}{arXiv preprint arXiv:1603.09025} .
\bibitem[{Donahue et~al.(2015)Donahue, Anne~Hendricks, Guadarrama, Rohrbach,
  Venugopalan and Saenko~et al.}]{donahue2015long}
\bibinfo{author}{Donahue, J.}, \bibinfo{author}{Anne~Hendricks, L.},
  \bibinfo{author}{Guadarrama, S.}, \bibinfo{author}{Rohrbach, M.},
  \bibinfo{author}{Venugopalan, S.}, \bibinfo{author}{Saenko~et al., K.},
  \bibinfo{year}{2015}.
\newblock \bibinfo{title}{Long-term recurrent convolutional networks for visual
  recognition and description}, in: \bibinfo{booktitle}{{Conference on Computer
  Vision and Pattern Recognition}}, pp. \bibinfo{pages}{2625--2634}.
\bibitem[{El~Sallab et~al.(2018)El~Sallab, Sobh, Zidan, Zahran and
  Abdelkarim}]{el2018yolo4d}
\bibinfo{author}{El~Sallab, A.}, \bibinfo{author}{Sobh, I.},
  \bibinfo{author}{Zidan, M.}, \bibinfo{author}{Zahran, M.},
  \bibinfo{author}{Abdelkarim, S.}, \bibinfo{year}{2018}.
\newblock \bibinfo{title}{Yolo4d: A spatio-temporal approach for real-time
  multi-object detection and classification from lidar point clouds}, in:
  \bibinfo{booktitle}{Conference on Neural Information Processing Systems}.
\bibitem[{Feichtenhofer et~al.(2016)Feichtenhofer, Pinz and
  Zisserman}]{feichtenhofer2016convolutional}
\bibinfo{author}{Feichtenhofer, C.}, \bibinfo{author}{Pinz, A.},
  \bibinfo{author}{Zisserman, A.}, \bibinfo{year}{2016}.
\newblock \bibinfo{title}{Convolutional two-stream network fusion for video
  action recognition}, in: \bibinfo{booktitle}{Conference on Computer Vision
  and Pattern Recognition}, pp. \bibinfo{pages}{1933--1941}.
\bibitem[{Funke et~al.(2019)Funke, Bodenstedt, Oehme, von Bechtolsheim, Weitz
  and Speidel}]{funke2019using}
\bibinfo{author}{Funke, I.}, \bibinfo{author}{Bodenstedt, S.},
  \bibinfo{author}{Oehme, F.}, \bibinfo{author}{von Bechtolsheim, F.},
  \bibinfo{author}{Weitz, J.}, \bibinfo{author}{Speidel, S.},
  \bibinfo{year}{2019}.
\newblock \bibinfo{title}{Using 3d convolutional neural networks to learn
  spatiotemporal features for automatic surgical gesture recognition in video},
  in: \bibinfo{booktitle}{International Conference on Medical Image Computing
  and Computer-Assisted Intervention}, \bibinfo{organization}{Springer}. pp.
  \bibinfo{pages}{467--475}.
\bibitem[{Gao et~al.(2018)Gao, Liu, Peven, Unberath and
  Reiter}]{gao2018learning}
\bibinfo{author}{Gao, C.}, \bibinfo{author}{Liu, X.}, \bibinfo{author}{Peven,
  M.}, \bibinfo{author}{Unberath, M.}, \bibinfo{author}{Reiter, A.},
  \bibinfo{year}{2018}.
\newblock \bibinfo{title}{Learning to see forces: Surgical force prediction
  with rgb-point cloud temporal convolutional networks}, in:
  \bibinfo{booktitle}{OR 2.0 Context-Aware Operating Theaters, Computer
  Assisted Robotic Endoscopy, Clinical Image-Based Procedures, and Skin Image
  Analysis}, pp. \bibinfo{pages}{118--127}.
\bibitem[{Gessert et~al.(2018a)Gessert, Beringhoff, Otte and
  Schlaefer}]{gessert2018force}
\bibinfo{author}{Gessert, N.}, \bibinfo{author}{Beringhoff, J.},
  \bibinfo{author}{Otte, C.}, \bibinfo{author}{Schlaefer, A.},
  \bibinfo{year}{2018}a.
\newblock \bibinfo{title}{{Force estimation from OCT volumes using 3D CNNs}}.
\newblock \bibinfo{journal}{International Journal of Computer Assisted
  Radiology and Surgery} \bibinfo{volume}{13}, \bibinfo{pages}{1073–--1082}.
\bibitem[{Gessert et~al.(2018b)Gessert, Priegnitz, Saathoff, Antoni, Meyer and
  Hamann~et al.}]{gessert2018needle}
\bibinfo{author}{Gessert, N.}, \bibinfo{author}{Priegnitz, T.},
  \bibinfo{author}{Saathoff, T.}, \bibinfo{author}{Antoni, S.T.},
  \bibinfo{author}{Meyer, D.}, \bibinfo{author}{Hamann~et al., M.F.},
  \bibinfo{year}{2018}b.
\newblock \bibinfo{title}{Needle tip force estimation using an oct fiber and a
  fused convgru-cnn architecture}, in: \bibinfo{booktitle}{International
  Conference on Medical Image Computing and Computer-Assisted Intervention},
  pp. \bibinfo{pages}{222--229}.
\bibitem[{Gessert et~al.(2019)Gessert, Priegnitz, Saathoff, Antoni, Meyer and
  Hamann~et al.}]{gessert2019spatio}
\bibinfo{author}{Gessert, N.}, \bibinfo{author}{Priegnitz, T.},
  \bibinfo{author}{Saathoff, T.}, \bibinfo{author}{Antoni, S.T.},
  \bibinfo{author}{Meyer, D.}, \bibinfo{author}{Hamann~et al., M.F.},
  \bibinfo{year}{2019}.
\newblock \bibinfo{title}{{Spatio-temporal deep learning models for tip force
  estimation during needle insertion}}.
\newblock \bibinfo{journal}{{International Journal of Computer Assisted
  Radiology and Surgery}} \bibinfo{volume}{14}, \bibinfo{pages}{1485--1493}.
\bibitem[{Gessert et~al.(2018c)Gessert, Schl{\"u}ter and
  Schlaefer}]{gessert2018deep}
\bibinfo{author}{Gessert, N.}, \bibinfo{author}{Schl{\"u}ter, M.},
  \bibinfo{author}{Schlaefer, A.}, \bibinfo{year}{2018}c.
\newblock \bibinfo{title}{A deep learning approach for pose estimation from
  volumetric oct data}.
\newblock \bibinfo{journal}{Medical Image Analysis} \bibinfo{volume}{46},
  \bibinfo{pages}{162--179}.
\bibitem[{Greminger and Nelson(2003)}]{greminger2003modeling}
\bibinfo{author}{Greminger, M.A.}, \bibinfo{author}{Nelson, B.J.},
  \bibinfo{year}{2003}.
\newblock \bibinfo{title}{Modeling elastic objects with neural networks for
  vision-based force measurement}, in: \bibinfo{booktitle}{International
  Conference on Intelligent Robots and Systems}, pp.
  \bibinfo{pages}{1278--1283}.
\bibitem[{Greminger and Nelson(2004)}]{greminger2004vision}
\bibinfo{author}{Greminger, M.A.}, \bibinfo{author}{Nelson, B.J.},
  \bibinfo{year}{2004}.
\newblock \bibinfo{title}{Vision-based force measurement}.
\newblock \bibinfo{journal}{IEEE Transactions on Pattern Analysis and Machine
  Intelligence} \bibinfo{volume}{26}, \bibinfo{pages}{290--298}.
\bibitem[{Gupta et~al.(1999)Gupta, Jensen and de~Juan}]{gupta1999surgical}
\bibinfo{author}{Gupta, P.K.}, \bibinfo{author}{Jensen, P.S.},
  \bibinfo{author}{de~Juan, E.}, \bibinfo{year}{1999}.
\newblock \bibinfo{title}{Surgical forces and tactile perception during retinal
  microsurgery}, in: \bibinfo{booktitle}{International conference on medical
  image computing and computer-assisted intervention},
  \bibinfo{organization}{Springer}. pp. \bibinfo{pages}{1218--1225}.
\bibitem[{Haidegger et~al.(2009)Haidegger, Beny{\'o}, Kov{\'a}cs and
  Beny{\'o}}]{haidegger2009force}
\bibinfo{author}{Haidegger, T.}, \bibinfo{author}{Beny{\'o}, B.},
  \bibinfo{author}{Kov{\'a}cs, L.}, \bibinfo{author}{Beny{\'o}, Z.},
  \bibinfo{year}{2009}.
\newblock \bibinfo{title}{Force sensing and force control for surgical robots},
  in: \bibinfo{booktitle}{7th IFAC Symposium on Modeling and Control in
  Biomedical Systems}, pp. \bibinfo{pages}{413--418}.
\bibitem[{Haouchine et~al.(2018)Haouchine, Kuang, Cotin and
  Yip}]{haouchine2018vision}
\bibinfo{author}{Haouchine, N.}, \bibinfo{author}{Kuang, W.},
  \bibinfo{author}{Cotin, S.}, \bibinfo{author}{Yip, M.C.},
  \bibinfo{year}{2018}.
\newblock \bibinfo{title}{{Vision-based Force Feedback Estimation for
  Robot-assisted Surgery using Instrument-constrained Biomechanical 3D Maps}}.
\newblock \bibinfo{journal}{IEEE Robotics and Automation Letters} .
\bibitem[{He et~al.(2016a)He, Zhang, Ren and Sun}]{he2016deep}
\bibinfo{author}{He, K.}, \bibinfo{author}{Zhang, X.}, \bibinfo{author}{Ren,
  S.}, \bibinfo{author}{Sun, J.}, \bibinfo{year}{2016}a.
\newblock \bibinfo{title}{Deep residual learning for image recognition}, in:
  \bibinfo{booktitle}{{Conference on Computer Vision and Pattern Recognition}},
  pp. \bibinfo{pages}{770--778}.
\bibitem[{He et~al.(2016b)He, Zhang, Ren and Sun}]{he2016identity}
\bibinfo{author}{He, K.}, \bibinfo{author}{Zhang, X.}, \bibinfo{author}{Ren,
  S.}, \bibinfo{author}{Sun, J.}, \bibinfo{year}{2016}b.
\newblock \bibinfo{title}{Identity mappings in deep residual networks}, in:
  \bibinfo{booktitle}{European Conference on Computer Vision}, pp.
  \bibinfo{pages}{630--645}.
\bibitem[{Hochreiter and Schmidhuber(1997)}]{hochreiter1997long}
\bibinfo{author}{Hochreiter, S.}, \bibinfo{author}{Schmidhuber, J.},
  \bibinfo{year}{1997}.
\newblock \bibinfo{title}{Long short-term memory}.
\newblock \bibinfo{journal}{Neural Computation} \bibinfo{volume}{9},
  \bibinfo{pages}{1735--1780}.
\bibitem[{Ji et~al.(2013)Ji, Xu, Yang and Yu}]{ji20133d}
\bibinfo{author}{Ji, S.}, \bibinfo{author}{Xu, W.}, \bibinfo{author}{Yang, M.},
  \bibinfo{author}{Yu, K.}, \bibinfo{year}{2013}.
\newblock \bibinfo{title}{{3D convolutional neural networks for human action
  recognition}}.
\newblock \bibinfo{journal}{IEEE Transactions on Pattern Analysis and Machine
  Intelligence} \bibinfo{volume}{35}, \bibinfo{pages}{221--231}.
\bibitem[{Jin et~al.(2019)Jin, Li, Dou, Chen, Qin, Fu and Heng}]{jin2019multi}
\bibinfo{author}{Jin, Y.}, \bibinfo{author}{Li, H.}, \bibinfo{author}{Dou, Q.},
  \bibinfo{author}{Chen, H.}, \bibinfo{author}{Qin, J.}, \bibinfo{author}{Fu,
  C.W.}, \bibinfo{author}{Heng, P.A.}, \bibinfo{year}{2019}.
\newblock \bibinfo{title}{Multi-task recurrent convolutional network with
  correlation loss for surgical video analysis}.
\newblock \bibinfo{journal}{Medical image analysis} , \bibinfo{pages}{101572}.
\bibitem[{Karimirad et~al.(2014)Karimirad, Chauhan and
  Shirinzadeh}]{karimirad2014vision}
\bibinfo{author}{Karimirad, F.}, \bibinfo{author}{Chauhan, S.},
  \bibinfo{author}{Shirinzadeh, B.}, \bibinfo{year}{2014}.
\newblock \bibinfo{title}{Vision-based force measurement using neural networks
  for biological cell microinjection}.
\newblock \bibinfo{journal}{Journal of Biomechanics} \bibinfo{volume}{47},
  \bibinfo{pages}{1157--1163}.
\bibitem[{Kim et~al.(2010)Kim, Janabi-Sharifi and Kim}]{kim2010haptic}
\bibinfo{author}{Kim, J.}, \bibinfo{author}{Janabi-Sharifi, F.},
  \bibinfo{author}{Kim, J.}, \bibinfo{year}{2010}.
\newblock \bibinfo{title}{A haptic interaction method using visual information
  and physically based modeling}.
\newblock \bibinfo{journal}{IEEE/ASME Transactions on Mechatronics}
  \bibinfo{volume}{15}, \bibinfo{pages}{636--645}.
\bibitem[{Kim et~al.(2012)Kim, Seung, Choi, Park, Ko and Park}]{kim2012image}
\bibinfo{author}{Kim, W.}, \bibinfo{author}{Seung, S.}, \bibinfo{author}{Choi,
  H.}, \bibinfo{author}{Park, S.}, \bibinfo{author}{Ko, S.Y.},
  \bibinfo{author}{Park, J.O.}, \bibinfo{year}{2012}.
\newblock \bibinfo{title}{Image-based force estimation of deformable tissue
  using depth map for single-port surgical robot}, in:
  \bibinfo{booktitle}{International Conference on Control, Automation and
  Systems}, \bibinfo{organization}{IEEE}. pp. \bibinfo{pages}{1716--1719}.
\bibitem[{Kingma and Ba(2015)}]{Kingma.2014}
\bibinfo{author}{Kingma, D.}, \bibinfo{author}{Ba, J.}, \bibinfo{year}{2015}.
\newblock \bibinfo{title}{{Adam: A method for stochastic optimization}}, in:
  \bibinfo{booktitle}{{International Conference on Learning Representations}}.
\bibitem[{Kroh and Chalikonda(2015)}]{kroh2015essentials}
\bibinfo{author}{Kroh, M.}, \bibinfo{author}{Chalikonda, S.},
  \bibinfo{year}{2015}.
\newblock \bibinfo{title}{Essentials of robotic surgery}.
\newblock \bibinfo{publisher}{Springer}.
\bibitem[{van~de Leemput et~al.(2019)van~de Leemput, Prokop, van Ginneken and
  Manniesing}]{van2019stacked}
\bibinfo{author}{van~de Leemput, S.C.}, \bibinfo{author}{Prokop, M.},
  \bibinfo{author}{van Ginneken, B.}, \bibinfo{author}{Manniesing, R.},
  \bibinfo{year}{2019}.
\newblock \bibinfo{title}{Stacked bidirectional convolutional lstms for
  deriving 3d non-contrast ct from spatiotemporal 4d ct}.
\newblock \bibinfo{journal}{IEEE transactions on medical imaging} .
\bibitem[{Liu et~al.(2016)Liu, Zhang and Tian}]{liu20163d}
\bibinfo{author}{Liu, Z.}, \bibinfo{author}{Zhang, C.}, \bibinfo{author}{Tian,
  Y.}, \bibinfo{year}{2016}.
\newblock \bibinfo{title}{{3D-based deep convolutional neural network for
  action recognition with depth sequences}}.
\newblock \bibinfo{journal}{Image and Vision Computing} \bibinfo{volume}{55},
  \bibinfo{pages}{93--100}.
\bibitem[{Marban et~al.(2019)Marban, Srinivasan, Samek, Fern{\'a}ndez and
  Casals}]{marban2019recurrent}
\bibinfo{author}{Marban, A.}, \bibinfo{author}{Srinivasan, V.},
  \bibinfo{author}{Samek, W.}, \bibinfo{author}{Fern{\'a}ndez, J.},
  \bibinfo{author}{Casals, A.}, \bibinfo{year}{2019}.
\newblock \bibinfo{title}{A recurrent convolutional neural network approach for
  sensorless force estimation in robotic surgery}.
\newblock \bibinfo{journal}{Biomedical Signal Processing and Control}
  \bibinfo{volume}{50}, \bibinfo{pages}{134--150}.
\bibitem[{Maturana and Scherer(2015)}]{maturana2015voxnet}
\bibinfo{author}{Maturana, D.}, \bibinfo{author}{Scherer, S.},
  \bibinfo{year}{2015}.
\newblock \bibinfo{title}{{Voxnet: A 3D convolutional neural network for
  real-time object recognition}}, in: \bibinfo{booktitle}{International
  Conference on Intelligent Robots and Systems}, pp. \bibinfo{pages}{922--928}.
\bibitem[{McCreery et~al.(2008)McCreery, Trejos, Naish, Patel and
  Malthaner}]{mccreery2008feasibility}
\bibinfo{author}{McCreery, G.L.}, \bibinfo{author}{Trejos, A.L.},
  \bibinfo{author}{Naish, M.D.}, \bibinfo{author}{Patel, R.V.},
  \bibinfo{author}{Malthaner, R.A.}, \bibinfo{year}{2008}.
\newblock \bibinfo{title}{Feasibility of locating tumours in lung via
  kinaesthetic feedback}.
\newblock \bibinfo{journal}{The International Journal of Medical Robotics and
  Computer Assisted Surgery} \bibinfo{volume}{4}, \bibinfo{pages}{58--68}.
\bibitem[{Mozaffari et~al.(2014)Mozaffari, Behzadipour and
  Kohani}]{mozaffari2014identifying}
\bibinfo{author}{Mozaffari, A.}, \bibinfo{author}{Behzadipour, S.},
  \bibinfo{author}{Kohani, M.}, \bibinfo{year}{2014}.
\newblock \bibinfo{title}{Identifying the tool-tissue force in robotic
  laparoscopic surgery using neuro-evolutionary fuzzy systems and a synchronous
  self-learning hyper level supervisor}.
\newblock \bibinfo{journal}{Applied Soft Computing} \bibinfo{volume}{14},
  \bibinfo{pages}{12--30}.
\bibitem[{Myronenko et~al.(2020)Myronenko, Yang, Buch, Xu, Ihsani and Doyle~et
  al.}]{myronenko20194d}
\bibinfo{author}{Myronenko, A.}, \bibinfo{author}{Yang, D.},
  \bibinfo{author}{Buch, V.}, \bibinfo{author}{Xu, D.},
  \bibinfo{author}{Ihsani, A.}, \bibinfo{author}{Doyle~et al., S.},
  \bibinfo{year}{2020}.
\newblock \bibinfo{title}{4d cnn for semantic segmentation of cardiac
  volumetric sequences}, in: \bibinfo{booktitle}{Statistical Atlases and
  Computational Models of the Heart. Multi-Sequence CMR Segmentation,
  CRT-EPiggy and LV Full Quantification Challenges},
  \bibinfo{publisher}{Springer International Publishing}. pp.
  \bibinfo{pages}{72--80}.
\bibitem[{Noohi et~al.(2014)Noohi, Parastegari and
  {\v{Z}}efran}]{noohi2014using}
\bibinfo{author}{Noohi, E.}, \bibinfo{author}{Parastegari, S.},
  \bibinfo{author}{{\v{Z}}efran, M.}, \bibinfo{year}{2014}.
\newblock \bibinfo{title}{Using monocular images to estimate interaction forces
  during minimally invasive surgery}, in: \bibinfo{booktitle}{International
  Conference on Intelligent Robots and Systems}, pp.
  \bibinfo{pages}{4297--4302}.
\bibitem[{Ord{\'o}{\~n}ez and Roggen(2016)}]{ordonez2016deep}
\bibinfo{author}{Ord{\'o}{\~n}ez, F.J.}, \bibinfo{author}{Roggen, D.},
  \bibinfo{year}{2016}.
\newblock \bibinfo{title}{Deep convolutional and lstm recurrent neural networks
  for multimodal wearable activity recognition}.
\newblock \bibinfo{journal}{Sensors} \bibinfo{volume}{16},
  \bibinfo{pages}{115}.
\bibitem[{Otte et~al.(2016)Otte, Beringhoff, Latus, Antoni, Rajput and
  Schlaefer}]{otte2016towards}
\bibinfo{author}{Otte, C.}, \bibinfo{author}{Beringhoff, J.},
  \bibinfo{author}{Latus, S.}, \bibinfo{author}{Antoni, S.T.},
  \bibinfo{author}{Rajput, O.}, \bibinfo{author}{Schlaefer, A.},
  \bibinfo{year}{2016}.
\newblock \bibinfo{title}{Towards force sensing based on instrument-tissue
  interaction}, in: \bibinfo{booktitle}{Multisensor Fusion and Integration for
  Intelligent Systems}, pp. \bibinfo{pages}{180--185}.
\bibitem[{Pigou et~al.(2018)Pigou, Van Den~Oord, Dieleman, Van~Herreweghe and
  Dambre}]{pigou2018beyond}
\bibinfo{author}{Pigou, L.}, \bibinfo{author}{Van Den~Oord, A.},
  \bibinfo{author}{Dieleman, S.}, \bibinfo{author}{Van~Herreweghe, M.},
  \bibinfo{author}{Dambre, J.}, \bibinfo{year}{2018}.
\newblock \bibinfo{title}{Beyond temporal pooling: Recurrence and temporal
  convolutions for gesture recognition in video}.
\newblock \bibinfo{journal}{International Journal of Computer Vision}
  \bibinfo{volume}{126}, \bibinfo{pages}{430--439}.
\bibitem[{Qi et~al.(2017)Qi, Su, Mo and Guibas}]{qi2017pointnet}
\bibinfo{author}{Qi, C.R.}, \bibinfo{author}{Su, H.}, \bibinfo{author}{Mo, K.},
  \bibinfo{author}{Guibas, L.J.}, \bibinfo{year}{2017}.
\newblock \bibinfo{title}{{Pointnet: Deep learning on point sets for 3D
  classification and segmentation}}, in: \bibinfo{booktitle}{{Conference on
  Computer Vision and Pattern Recognition}}.
\bibitem[{Qiu et~al.(2017)Qiu, Yao and Mei}]{qiu2017learning}
\bibinfo{author}{Qiu, Z.}, \bibinfo{author}{Yao, T.}, \bibinfo{author}{Mei,
  T.}, \bibinfo{year}{2017}.
\newblock \bibinfo{title}{{Learning spatio-temporal representation with
  pseudo-3D residual networks}}, in: \bibinfo{booktitle}{International
  Conference on Computer Vision}, pp. \bibinfo{pages}{5534--5542}.
\bibitem[{Siddiqui et~al.(2018)Siddiqui, Nam, Tozburun, Lippok, Blatter and
  Vakoc}]{siddiqui2018high}
\bibinfo{author}{Siddiqui, M.}, \bibinfo{author}{Nam, A.S.},
  \bibinfo{author}{Tozburun, S.}, \bibinfo{author}{Lippok, N.},
  \bibinfo{author}{Blatter, C.}, \bibinfo{author}{Vakoc, B.J.},
  \bibinfo{year}{2018}.
\newblock \bibinfo{title}{High-speed optical coherence tomography by circular
  interferometric ranging}.
\newblock \bibinfo{journal}{Nature Photonics} \bibinfo{volume}{12},
  \bibinfo{pages}{111}.
\bibitem[{Simonyan and Zisserman(2014)}]{simonyan2014two}
\bibinfo{author}{Simonyan, K.}, \bibinfo{author}{Zisserman, A.},
  \bibinfo{year}{2014}.
\newblock \bibinfo{title}{Two-stream convolutional networks for action
  recognition in videos}, in: \bibinfo{booktitle}{Advances in Neural
  Information Processing Systems}, pp. \bibinfo{pages}{568--576}.
\bibitem[{Sokhanvar et~al.(2012)Sokhanvar, Dargahi, Najarian and
  Arbatani}]{sokhanvar2012clinical}
\bibinfo{author}{Sokhanvar, S.}, \bibinfo{author}{Dargahi, J.},
  \bibinfo{author}{Najarian, S.}, \bibinfo{author}{Arbatani, S.},
  \bibinfo{year}{2012}.
\newblock \bibinfo{title}{Clinical and regulatory challenges for medical
  devices tactile sensing and displays}.
\newblock \bibinfo{journal}{Haptic Feedback for Minimally Invasive Surgery and
  Robotics} .
\bibitem[{Sun et~al.(2015)Sun, Jia, Yeung and Shi}]{sun2015human}
\bibinfo{author}{Sun, L.}, \bibinfo{author}{Jia, K.}, \bibinfo{author}{Yeung,
  D.Y.}, \bibinfo{author}{Shi, B.E.}, \bibinfo{year}{2015}.
\newblock \bibinfo{title}{Human action recognition using factorized
  spatio-temporal convolutional networks}, in:
  \bibinfo{booktitle}{International Conference on Computer Vision}, pp.
  \bibinfo{pages}{4597--4605}.
\bibitem[{Tran et~al.(2015)Tran, Bourdev, Fergus, Torresani and
  Paluri}]{tran2015learning}
\bibinfo{author}{Tran, D.}, \bibinfo{author}{Bourdev, L.},
  \bibinfo{author}{Fergus, R.}, \bibinfo{author}{Torresani, L.},
  \bibinfo{author}{Paluri, M.}, \bibinfo{year}{2015}.
\newblock \bibinfo{title}{{Learning spatiotemporal features with 3D
  convolutional networks}}, in: \bibinfo{booktitle}{International Conference on
  Computer Vision}, pp. \bibinfo{pages}{4489--4497}.
\bibitem[{Tran et~al.(2018)Tran, Wang, Torresani, Ray, LeCun and
  Paluri}]{tran2018closer}
\bibinfo{author}{Tran, D.}, \bibinfo{author}{Wang, H.},
  \bibinfo{author}{Torresani, L.}, \bibinfo{author}{Ray, J.},
  \bibinfo{author}{LeCun, Y.}, \bibinfo{author}{Paluri, M.},
  \bibinfo{year}{2018}.
\newblock \bibinfo{title}{A closer look at spatiotemporal convolutions for
  action recognition}, in: \bibinfo{booktitle}{Conference on Computer Vision
  and Pattern Recognition}, pp. \bibinfo{pages}{6450--6459}.
\bibitem[{Trejos et~al.(2010)Trejos, Patel and Naish}]{trejos2010force}
\bibinfo{author}{Trejos, A.}, \bibinfo{author}{Patel, R.},
  \bibinfo{author}{Naish, M.}, \bibinfo{year}{2010}.
\newblock \bibinfo{title}{Force sensing and its application in minimally
  invasive surgery and therapy: a survey}.
\newblock \bibinfo{journal}{Proceedings of the Institution of Mechanical
  Engineers, Part C: Journal of Mechanical Engineering Science}
  \bibinfo{volume}{224}, \bibinfo{pages}{1435--1454}.
\bibitem[{Varol et~al.(2018)Varol, Laptev and Schmid}]{varol2018long}
\bibinfo{author}{Varol, G.}, \bibinfo{author}{Laptev, I.},
  \bibinfo{author}{Schmid, C.}, \bibinfo{year}{2018}.
\newblock \bibinfo{title}{Long-term temporal convolutions for action
  recognition}.
\newblock \bibinfo{journal}{IEEE Transactions on Pattern Analysis and Machine
  Intelligence} \bibinfo{volume}{40}, \bibinfo{pages}{1510--1517}.
\bibitem[{Wang et~al.(2016)Wang, Xiong, Wang, Qiao, Lin, Tang and
  Van~Gool}]{wang2016temporal}
\bibinfo{author}{Wang, L.}, \bibinfo{author}{Xiong, Y.}, \bibinfo{author}{Wang,
  Z.}, \bibinfo{author}{Qiao, Y.}, \bibinfo{author}{Lin, D.},
  \bibinfo{author}{Tang, X.}, \bibinfo{author}{Van~Gool, L.},
  \bibinfo{year}{2016}.
\newblock \bibinfo{title}{Temporal segment networks: Towards good practices for
  deep action recognition}, in: \bibinfo{booktitle}{European Conference on
  Computer Vision}, pp. \bibinfo{pages}{20--36}.
\bibitem[{Xingjian et~al.(2015)Xingjian, Chen, Wang, Yeung, Wong and
  Woo}]{xingjian2015convolutional}
\bibinfo{author}{Xingjian, S.}, \bibinfo{author}{Chen, Z.},
  \bibinfo{author}{Wang, H.}, \bibinfo{author}{Yeung, D.Y.},
  \bibinfo{author}{Wong, W.K.}, \bibinfo{author}{Woo, W.c.},
  \bibinfo{year}{2015}.
\newblock \bibinfo{title}{{Convolutional LSTM network: A machine learning
  approach for precipitation nowcasting}}, in: \bibinfo{booktitle}{Advances in
  Neural Information Processing Systems}, pp. \bibinfo{pages}{802--810}.
\bibitem[{Yue-Hei~Ng et~al.(2015)Yue-Hei~Ng, Hausknecht, Vijayanarasimhan,
  Vinyals, Monga and Toderici}]{yue2015beyond}
\bibinfo{author}{Yue-Hei~Ng, J.}, \bibinfo{author}{Hausknecht, M.},
  \bibinfo{author}{Vijayanarasimhan, S.}, \bibinfo{author}{Vinyals, O.},
  \bibinfo{author}{Monga, R.}, \bibinfo{author}{Toderici, G.},
  \bibinfo{year}{2015}.
\newblock \bibinfo{title}{Beyond short snippets: Deep networks for video
  classification}, in: \bibinfo{booktitle}{{Conference on Computer Vision and
  Pattern Recognition}}, pp. \bibinfo{pages}{4694--4702}.
\bibitem[{Zhao et~al.(2018)Zhao, Li, Zhang, Zhao, Makkie, Zhang, Li and
  Liu}]{zhao2018modeling}
\bibinfo{author}{Zhao, Y.}, \bibinfo{author}{Li, X.}, \bibinfo{author}{Zhang,
  W.}, \bibinfo{author}{Zhao, S.}, \bibinfo{author}{Makkie, M.},
  \bibinfo{author}{Zhang, M.}, \bibinfo{author}{Li, Q.}, \bibinfo{author}{Liu,
  T.}, \bibinfo{year}{2018}.
\newblock \bibinfo{title}{Modeling 4d fmri data via spatio-temporal
  convolutional neural networks (st-cnn)}, in:
  \bibinfo{booktitle}{International Conference on Medical Image Computing and
  Computer-Assisted Intervention}, \bibinfo{organization}{Springer}. pp.
  \bibinfo{pages}{181--189}.

\end{thebibliography}



\end{document}